\documentclass[runningheads]{llncs}

\usepackage[mobile]{eccv}

\usepackage{eccvabbrv}

\usepackage{graphicx}
\usepackage{booktabs}

\usepackage[accsupp]{axessibility}  

\usepackage{multirow} 
\usepackage{colortbl}  
\definecolor{commentcolor}{HTML}{93a1a1}
\usepackage{algorithm} 
\usepackage{algpseudocode} 
\usepackage{xcolor}
\definecolor{graybg}{gray}{0.9}
\definecolor{lightred}{RGB}{255, 102, 102}

\usepackage{amsmath}
\usepackage{tabularx} 

\usepackage{hyperref}

\usepackage{orcidlink}
\newcommand{\ourmethod}[1]{\textsc{GIDE}}
\newcommand{\ourbench}[1]{GIDE-Bench}
\usepackage{wrapfig}

\begin{document}

\title{\ourmethod{}: Unlocking Diffusion LLMs for Precise Training-Free Image Editing} 

\titlerunning{\ourmethod{}: Unlocking DLLMs for Training-Free Image Editing}

\author{Zifeng Zhu\inst{1}\orcidlink{0009-0002-6330-8355} \and
Jiaming Han\inst{2}\orcidlink{0000-0002-3325-4520} \and
Jiaxiang Zhao\inst{1} \and
Minnan Luo\inst{1}\orcidlink{0000-0002-0140-7860} \and \\
Xiangyu Yue\inst{2}\orcidlink{0000-0002-6887-2046}
}
\authorrunning{Z.~Zhu et al.}

\institute{Xi'an Jiaotong University \and
MMLab, The Chinese University of Hong Kong \\
\email{zivenzhu@stu.xjtu.edu.cn}}

\maketitle

\begin{abstract}

While Diffusion Large Language Models (DLLMs) have demonstrated remarkable capabilities in multi-modal generation, performing precise, training-free image editing remains an open challenge. Unlike continuous diffusion models, the discrete tokenization inherent in DLLMs hinders the application of standard noise inversion techniques, often leading to structural degradation during editing. In this paper, we introduce \textbf{GIDE} (Grounded Inversion for DLLM Image Editing), a unified framework designed to bridge this gap. GIDE incorporates a novel \textbf{Discrete Noise Inversion} mechanism that accurately captures latent noise patterns within the discrete token space, ensuring high-fidelity reconstruction. We then decompose the editing pipeline into \textbf{grounding}, \textbf{inversion}, and \textbf{refinement} stages. This design enables \ourmethod{} supporting various editing instructions (text, point and box) and operations while strictly preserving the unedited background. Furthermore, to overcome the limitations of existing single-step evaluation protocols, we introduce \ourbench{}, a rigorous benchmark comprising 805 compositional editing scenarios guided by diverse multi-modal inputs. Extensive experiments on \ourbench{} demonstrate that \ourmethod{} significantly outperforms prior training-free methods, improving Semantic Correctness by 51.83\% and Perceptual Quality by 50.39\%. Additional evaluations on ImgEdit-Bench confirm its broad applicability, demonstrating consistent gains over trained baselines and yielding photorealistic consistency on par with leading models.\footnote{Data and code are available at \href{https://github.com/Zivenzhu/GIDE}{https://github.com/Zivenzhu/GIDE}.}.

\end{abstract}

\section{Introduction}
\label{sec:intro}

\begin{figure}[tb]
  \centering
  \includegraphics[width=\linewidth]{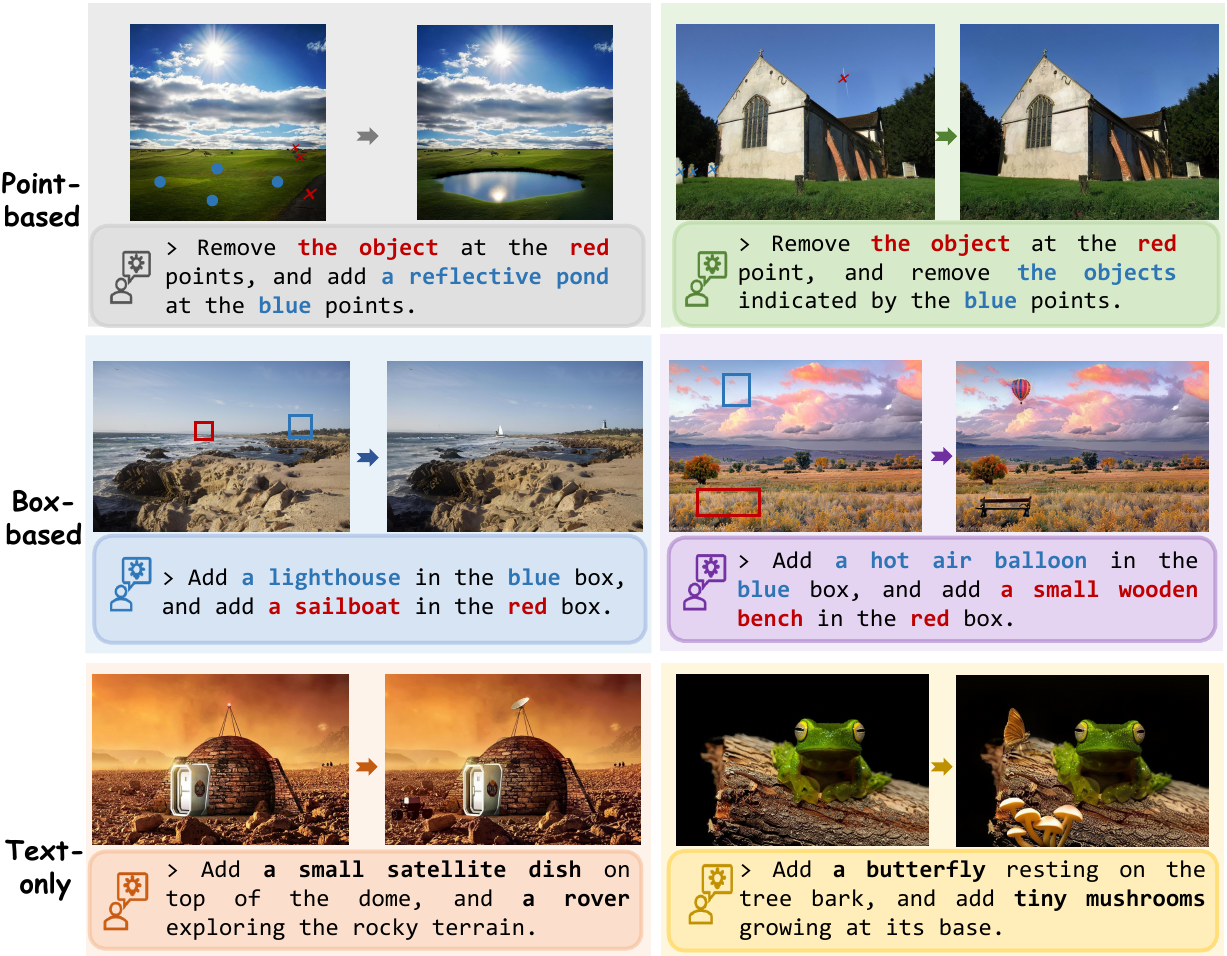}
    \caption{\textbf{Demonstration of our proposed \ourmethod{} framework}. Given a source image (left of each pair) and an editing instruction guided by flexible spatial modalities (points, bounding boxes, or pure text), \ourmethod{} enables precise and localized image editing. The edited results (right of each pair) demonstrate high-quality visual synthesis and strict background preservation, all achieved in a completely training-free manner.}
    \vspace{-3mm}
  \label{fig:intro}
\end{figure}

Diffusion Large Language Models (DLLMs)~\cite{liang2025discrete, liu2025longllada, shi2025muddit} have emerged as a powerful paradigm for unified multimodal modeling. Unlike autoregressive models constrained by sequential dependency~\cite{liu2024lumina, chen2025janus, xin2025resurrect, team2024chameleon}, DLLMs achieves superior sampling efficiency through parallel token generation~\cite{xin2025lumina, tian2025mmada}. Despite their potential, the application of DLLMs to image editing remains largely under-explored. 

Existing editing approaches faces significant challenge due to the inherent discrete nature of DLLMs~\cite{chang2022maskgit, nielarge, yang2025mmada}. On the one hand, finetuning-based methods often struggle to balance editability and fidelity~\cite{xin2025lumina}. On the other hand, training-free methods are currently hindered by the lack of a principled inversion mechanism tailored for discrete token spaces. Unlike standard diffusion models where continuous noise inversion (e.g., DDIM inversion~\cite{dhariwal2021diffusion, songdenoising}) is well-established, the discrete, LLM-based diffusion process lacks a counterpart to accurately map images back to their initial noise states. This absence leads to poor semantic preservation and severe visual artifacts, ultimately limiting the practical utility of DLLMs for precise editing tasks and hindering their widespread adoption.

To bridge this critical gap, we present \textbf{Grounded Inversion for DLLM Image Editing (\ourmethod{})}, the first training-free framework specifically designed for precise inversion and editing within the discrete token space of DLLMs. Operating entirely at inference time, \ourmethod{} adapts flexibly to diverse editing scenarios (see Fig.~\ref{fig:intro}) without the need for additional data annotation or model retraining. The core philosophy of \ourmethod{} is to decompose the complex editing task into three synergistic stages, each addressing a distinct sub-problem:
\begin{itemize}
    \item \textbf{Grounding for Localization}. To determine \textit{where} to edit, we introduce a robust grounding module that accepts flexible user inputs, ranging from points to boxes to text descriptions, to precisely localize target regions, ensuring modifications are strictly confined to the intended area.
    \item \textbf{Discrete Inversion for Preservation}. To determine \textit{how} to edit while preserving the original context, we propose a novel inversion mechanism tailored for discrete DLLMs. This cornerstone module projects the source image to a structure-aware latent representation, establishing a stable foundation for high-fidelity editing in complex visual scenarios.
    \item \textbf{Refinement for Harmonization}. Finally, to ensure visual coherence, a refinement module post-processes the edited regions, seamlessly blending them with the background preservation compared to prior attempted but also enables versatile editing capabilities across various modalities.
\end{itemize}

Beyond our methodological contributions, we identify a critical gap in current evaluation protocols. Existing benchmarks primarily focus on single-step text instructions~\cite{sheynin2024emu, ge2024seed, zhao2024ultraedit, yu2025anyedit}, failing to capture the complexity of compositional editing and spatial control that \ourmethod{} aim to solve. Furthermore, standard metrics like CLIPScore~\cite{hessel2021clipscore} often favor source preservation over editing faithfulness~\cite{krojer2024learning, qian2025gie}, allowing models to ``cheat'' by ignoring instructions. To rigorously validate \ourmethod{} against these higher standards, we introduce \textbf{\ourbench{}}, a challenging benchmark comprising 805 compositional editing scenarios guided by diverse multimodal inputs. Unlike prior datasets, \ourbench{} emphasizes precise region control and multi-step reasoning. We couple this with a holistic evaluation protocol that combines dual-model (GPT and Gemini) assessments for semantic correctness and perceptual quality with strict masked-based metrics for background preservation, ensuring a faithful analysis of whether models can modify intended content while strictly preserving the surrounding context.

Extensive experiments on \ourbench{} demonstrate that \ourmethod{} significantly outperforms state-of-the-art training-free methods, improving semantic correctness by 51.83\% and perceptual quality by 50.39\%. Notably, it achieves photorealistic consistency comparable to leading models, effectively preserving depth and lighting interfaces often lost in prior works. Further evaluation on ImgEdit-Bench~\cite{ye2025imgedit} confirm its robustness, surpassing the trained baseline model by up to 39.13\% on complex editing tasks. In summary, our work makes \textbf{three key contributions}: \textbf{1)} The first discrete noise inversion mechanism tailored for DLLMs. \textbf{2)} A unified, training-free editing framework supporting diverse editing prompts and scenarios. \textbf{3)} A rigorous benchmark for compositional editing evaluation.


\section{Related Work}

\subsection{Training-Free Image Editing}

Training-free image editing enables controllable modifications without additional training~\cite{nguyen2025h, fu2025feededit, chung2024style, morita2025tkg, avrahami2025stable, xu2025stylessp, mo2024freecontrol, zhu2025training, kim2025reflex, zhu2025kv, hu2025anchor}. These methods primarily fall into two paradigms: attention control and inversion. \textbf{Attention control methods}, mainly designed for diffusion models~\cite{rombach2022high}, manipulate attention maps to maintain consistency. For instance, Prompt-to-Prompt~\cite{hertzprompt} reuses source cross-attention maps during target generation to preserve backgrounds. MasaCtrl~\cite{cao2023masactrl} refines this for region-aware control by substituting key and value matrices in deeper layers and later timesteps. Recently, Add-it~\cite{teweladd} enables flexible editing by concatenating and dynamically weighting source and target attention representations. \textbf{Inversion-based methods} reconstruct an image's generation trajectory for faithful editing. DDIM Inversion~\cite{dhariwal2021diffusion, songdenoising} reverses the ODE process in diffusion models, while Null-text Inversion~\cite{mokady2023null} extends this to text-guided generation by optimizing the null-text embedding. Direct Inversion~\cite{jupnp} bypasses this optimization using logit differences. Furthermore, DICE~\cite{hedice} and VARIN~\cite{dao2025discrete} adapt inversion techniques to masked generative models and visual autoregressive models~\cite{chang2023muse, tian2024visual, tanghart}, respectively. Despite these advances, a principled inversion algorithm tailored for the inherently stochastic and discrete nature of DLLMs remains lacking, and methods like DICE exhibit notable limitations~(\cref{5.2.2:Preservation of Non-edited Regions.}). 

\subsection{Evaluation of Image Editing}

Early image editing benchmarks primarily rely on CLIP-based metrics to evaluate semantic alignment~\cite{ghosh2023geneval, sheynin2024emu, ruiz2023dreambooth, li2023dreamedit}, together with pixel-level metrics such as PSNR, SSIM, and MSE to assess the preservation of non-edited regions, as exemplified by PIE-Bench~\cite{jupnp} and MagicBrush~\cite{zhang2023magicbrush}. However, CLIP-based evaluation has been shown to be unreliable, since trivial solutions such as directly copying the original image may still achieve high CLIP scores due to their inability to disentangle fine-grained local modifications from global context~\cite{krojer2024learning}. With the rapid advancement of vision-language models (VLMs), recent benchmarks have increasingly adopted VLM-based evaluators~\cite{pan2025ice, pathiraja2025refedit, gumulti, wang2025gpt, pengdreambench++}, for example GPT-4o~\cite{gpt4o}, which demonstrate substantially higher agreement with human judgments~\cite{ye2025imgedit}. Representative examples include ImgEdit-Bench~\cite{ye2025imgedit}, GEdit-Bench~\cite{liu2025step1x}, and Omni-Edit-Bench~\cite{wei2024omniedit}. Nevertheless, relying solely on VLMs is insufficient, as they struggle to capture fine-grained structural distortions and quantify background preservation at the pixel level. GIE-Bench~\cite{qian2025gie} further combines pixel-level metrics with GPT-4o-based evaluation, but it relies on predefined edited regions, which may not accurately reflect the actual edited areas. To address this, \ourbench{} dynamically grounds edited regions and incorporates multi-modal instructions for a more challenging and comprehensive evaluation.


\section{Approach}

\begin{figure}[tb]
  \centering
  \includegraphics[width=\linewidth]{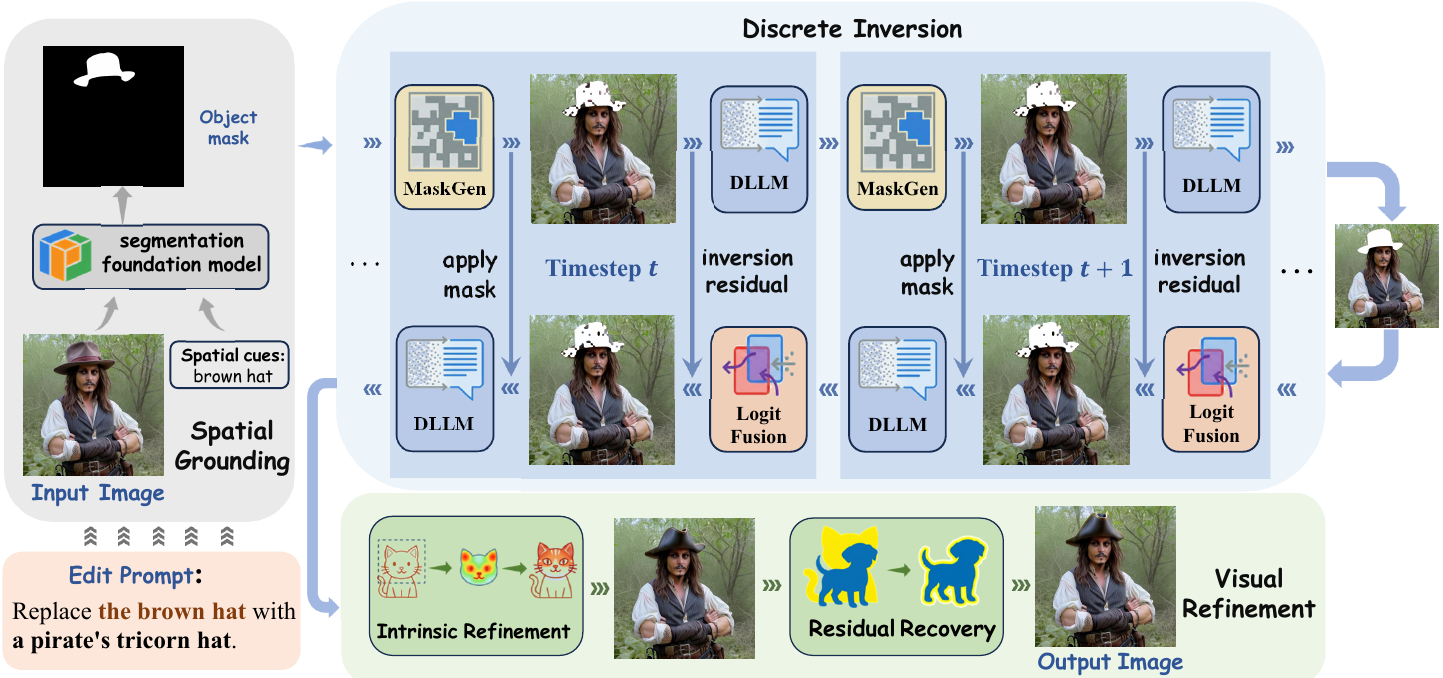}
    \caption{\textbf{The overall architecture of \ourmethod{}}. It decouples the editing process into three sequential stages: \textbf{Grounding} locates the target region via a segmentation foundation model; \textbf{Inversion} executes the edit by reconstructing the discrete latent space; and \textbf{Refinement} enhances the visual coherence and fidelity of the final output.}
    \vspace{-3mm}
  \label{fig:method}
\end{figure}

In this section, we introduce \textbf{Grounded Inversion for DLLM Image Editing (\ourmethod{})}, a generalized framework for training-free image editing using DLLMs. To overcome the limitations of existing methods in discrete latent space, \ourmethod{} adopts a modular design comprising \textbf{Grounding}, \textbf{Inversion}, and \textbf{Refinement} (see Fig.~\ref{fig:method}). This architecture allows us to decouple the ``where'' (location via grounding) from the ``how'' (reconstruction via discrete inversion) and the ``quality'' (enhancement via refinement). Consequently, our framework naturally extends to a wide range of editing operations without requiring task-specific tuning. In the following section, we will introduce each component in detail.

\subsection{Grounding-Aware Discrete Inversion}
\label{sec:inversion_module}

\begin{algorithm}[t]
\caption{\textbf{Grounding-Aware Inversion and Editing for DLLMs}}
\label{algorithm:inversion_diffusion_llm}
\begin{algorithmic}[1]

\Require
Input image $\boldsymbol{x}_0$; 
Grounding mask $\mathbf{M}$;
Total timesteps $T$; 
Mask token id $k_{\text{mask}}$; 
DLLM $\mathcal{D}_\theta$; 
Source prompt $\boldsymbol{c}$, Target prompt $\boldsymbol{c}'$; 
Mixing coef. $\lambda$

\Ensure
Edited image $\tilde{\boldsymbol{x}}_1$

\State \textcolor{gray}{\textit{// Stage 1: Inversion (Forward Process)}}
\For{$t = 1$ \textbf{to} $T$}
    \State $n_t \gets \texttt{SineSchedule}(t, T)$ 
    \Comment{Calculate mask ratio via sine function}
    \State $\boldsymbol{m}_t \gets \texttt{MaskGen}(\boldsymbol{x}_0, \mathbf{M}, n_t)$ 
    \Comment{Generate mask within $\mathbf{M}$}
    
    \State $\boldsymbol{x}_t \gets \boldsymbol{x}_0 \odot (\mathbf{1} - \boldsymbol{m}_t) + k_{\text{mask}} \cdot \boldsymbol{m}_t$
    \Comment{Apply mask tokens}
    
    \State $\hat{\boldsymbol{y}}_t \gets \mathcal{D}_\theta(\boldsymbol{x}_t, \boldsymbol{c}, t)$
    \Comment{Predict token logits}
    
    \State $\boldsymbol{y}_t \gets \texttt{LAI}(\boldsymbol{x}_t, \hat{\boldsymbol{y}}_t, \boldsymbol{x}_0)$
    \Comment{Construct ground-truth logits}
    
    \State $\boldsymbol{z}_t \gets \boldsymbol{y}_t - \hat{\boldsymbol{y}}_t$
    \Comment{Store inversion residual}
\EndFor

\State \textcolor{gray}{\textit{// Stage 2: Editing (Reverse Process)}}
\State $\tilde{\boldsymbol{x}}_{T+1} \gets \boldsymbol{x}_0$

\For{$t = T$ \textbf{to} $1$}
    \State $\boldsymbol{x}_t \gets \tilde{\boldsymbol{x}}_{t+1} \odot (\mathbf{1} - \boldsymbol{m}_t) + k_{\text{mask}} \cdot \boldsymbol{m}_t$
    \Comment{Re-apply grounding mask}
    
    \State $\hat{\boldsymbol{y}}_t \gets \mathcal{D}_\theta(\boldsymbol{x}_t, \boldsymbol{c}', t)$
    \Comment{Predict logits under target prompt}
    
    \State $\boldsymbol{g} \sim \text{Gumbel}(\mathbf{0}, \mathbf{I})$
    \Comment{Sample Gumbel noise}
    
    \State $\tilde{\boldsymbol{y}}_t \gets \hat{\boldsymbol{y}}_t + \lambda \cdot \boldsymbol{z}_t + (1-\lambda) \cdot \boldsymbol{g}$
    \Comment{Logit fusion with stochasticity}
    
    \State $\tilde{\boldsymbol{x}}_t \gets \arg\max \tilde{\boldsymbol{y}}_t$
    \Comment{Token selection}
\EndFor

\State \Return $\tilde{\boldsymbol{x}}_1$

\end{algorithmic}
\end{algorithm}


Different from continuous diffusion models that use deterministic ODEs for inversion, DLLMs rely on discrete, stochastic sampling. This means simply reversing the generation process is impossible: the same image could be generated by millions of different random paths. To solve this, we propose a \textbf{Grounding-Aware Discrete Inversion} algorithm (Algorithm~\ref{algorithm:inversion_diffusion_llm}). The core idea is to record the ``errors'' (residuals) the model makes when reconstructing the original image, and then replay these errors during editing to force the model to stay close to the original structure and prevent unintended visual deviations.


\noindent\textbf{Grounding-Constrained Masking.}\footnote{Implementation details of the grounding module are provided in Sec.~\ref{sec3.1:grounding_module}.} To prevent edits from leaking into the background, we must strictly control where the model is allowed to generate new tokens. We employ a sinusoidal masking schedule to determine the number of tokens to mask at step $t$: $n_t = \lfloor N \cdot \sin(\frac{\pi t}{2T}) \rfloor$, where $N$ is the total token count inside the grounding mask $\mathbf{M}$. This explicitly reverses the natural generation process: masking more tokens early on to capture fine-grained texture changes, and fewer tokens later to maintain global structural stability. To determine \emph{which} tokens to mask, we calculate a confidence score $s^{(i)}$ for each token $i$ based on the model's prediction probability. Crucially, to ensure progressive and strictly local masking, we enforce two constraints: 1) scores for tokens already masked in previous steps are set to $+\infty$, and 2) scores for tokens outside $\mathbf{M}$ are set to $-\infty$. The resulting step-specific binary mask $\boldsymbol{m}_t$ is then generated as:
\begin{equation*}
m_{t}^{(i)} =
\begin{cases}
1, & \text{if } s^{(i)} \in \text{top-}n_t(\mathbf{S}), \\
0, & \text{otherwise}.
\end{cases}
\end{equation*}
This ensures that the background $\boldsymbol{x}_{\text{bg}} = \boldsymbol{x}_0 \odot (\mathbf{1}-\mathbf{M})$ remains strictly preserved throughout the process, preventing any unintended visual alterations.


\noindent\textbf{Stochastic Logit Rectification.} To faithfully preserve the source object's appearance, we employ location-aware argmax inversion (LAI)~\cite{dao2025discrete} to extract a residual term $\boldsymbol{z}_t$, which encodes the structural priors of the original image. However, directly injecting this deterministic residual often restricts the generation process, leading to rigid artifacts and over-smoothed textures. To mitigate this, we introduce a stochastic fusion strategy that dynamically interpolates between the target semantics and source structure. Specifically, we rectify the target logits $\hat{\boldsymbol{y}}_t$ by incorporating the inversion residual $\boldsymbol{z}_t$ and a Gumbel noise term  $\boldsymbol{g}$, controlled by a mixing coefficient $\lambda$:
\begin{equation*}
    \tilde{\boldsymbol{y}}_t = \hat{\boldsymbol{y}}_t + \lambda \cdot \boldsymbol{z}_t + (1-\lambda) \cdot \boldsymbol{g}.
\end{equation*}
This formulation effectively balances the trade-off between semantic editability (driven by $\hat{\boldsymbol{y}}_t$) and structural fidelity (anchored by $\boldsymbol{z}_t$). Crucially, the controlled injection of Gumbel noise prevents the sampling distribution from collapsing into undesirable deterministic modes, thereby preserving fine-grained high-frequency details and ensuring a more natural synthesis.

\subsection{Multimodal Spatial Grounding}
\label{sec3.1:grounding_module}

To ensure high-fidelity preservation of non-edited regions, \ourmethod{} operates under a strict region-based constraint. Formally, given an input image $\mathbf{I} \in \mathbb{R}^{H \times W \times 3}$ and a user editing instruction $\mathcal{T}$, our goal is to derive a binary grounding mask $\mathbf{M} \in \{0, 1\}^{H \times W}$, where $\mathbf{M}_{i,j}=1$ indicates the editable foreground and $0$ represents the preserved background to guide the subsequent generation steps.

\noindent\textbf{Prompt-Driven Segmentation.} We harness the zero-shot generalization capabilities of state-of-the-art segmentation foundation models (SFMs)~\cite{carion2025sam, ravisam} to implement the grounding function $\mathcal{G}$. Depending on the granularity of the user input, the grounding process is formulated as $\mathbf{M} = \mathcal{G}(\mathbf{I}, \mathcal{P})$, where $\mathcal{P}$ represents the spatial cues derived from the instruction. For explicit spatial inputs (e.g., bounding boxes or points) or descriptive text prompts, we leverage a unified segmentation backbone capable of multi-modal prompting. This allows for direct extraction of object masks $\mathbf{M}$ with pixel-level precision, establishing a solid boundary for subsequent editing operations across various image domains.

\noindent\textbf{Attention-Guided Refinement.} To mitigate the brittleness of text-only grounding, we employ a fallback mechanism leveraging the DLLM's internal semantic knowledge. We compute a global heatmap $\mathbf{H} = \frac{1}{LK} \sum_{l,k} \mathbf{A}^{(l,k)}$ by averaging cross-attention maps $\mathbf{A}^{(l,k)}$ between visual tokens and text tokens across all layers and heads. High-activation points $\mathcal{P}_{attn} = \{ (x,y) \mid \mathbf{H}_{x,y} \in \text{top-}k(\mathbf{H}) \}$ are then extracted~\cite{teweladd} and utilized as foreground prompts for the segmentation model, ensuring robust mask generation in complex scenarios.

\subsection{High-Fidelity Visual Refinement}
\label{sec3.3:refinement_module}

Building upon the high semantic fidelity established by the inversion stage, we introduce a unified refinement module designed to further elevate visual coherence and boundary alignment. We formulate this refinement as a dual-stage process encompassing \textbf{intrinsic refinement} and \textbf{residual recovery}, formalized through set-theoretic operations on region masks. Specifically, let $\mathbf{M}_{\text{src}}$ denote the grounding mask of the original object (the region to be modified) and $\mathbf{M}_{\text{tgt}}$ denote the mask of the newly generated entity. By systematically manipulating the interplay between these regions, our module resolves potential texture inconsistencies and ensures seamless integration of the edited content into the background context.

\noindent\textbf{Intrinsic Refinement.} To resolve potential low-fidelity textures within generated regions, we introduce an uncertainty-aware refinement mechanism. We compute the confidence map $\mathbf{C}$ of the edited image and identify unstable tokens $\mathcal{U} = \{ (i,j) \mid \mathbf{C}_{i,j} < \tau \}$ with threshold $\tau$. The refinement mask is then defined as the intersection of unstable regions and the target: $\mathbf{M}_{\text{conf}} = \mathbf{M}_{\mathcal{U}} \cap \mathbf{M}_{\text{tgt}}$. We subsequently perform a localized re-sampling within $\mathbf{M}_{\text{conf}}$ to correct visual artifacts while keeping high-confidence structures intact.

\noindent\textbf{Residual Recovery.} To address shape mismatches between source and target objects, we employ a background recovery mechanism operating on the \emph{residual region}. Defined as the set difference $\mathbf{M}_{\text{res}} = \mathbf{M}_{\text{src}} \setminus \mathbf{M}_{\text{tgt}}$, this region isolates the specific area requiring context restoration. In \emph{Replace} and \emph{Remove} scenarios, $\mathbf{M}_{\text{res}}$ captures the exposed background gap for inpainting, whereas in \emph{Add}, it delineates the blending boundary with the original image. This consistent treatment of residuals ensures seamless integration across diverse editing modes.

\section{\ourbench{}}





\begin{figure}[htbp]
  \centering
  \begin{minipage}[c]{0.35\textwidth} 
    \centering
    \small
    \newcolumntype{L}{>{\raggedright\arraybackslash}X}
    \begin{tabularx}{\textwidth}{Lc} 
      \toprule
      \textbf{Category} & \textbf{Value} \\
      \midrule
      Total Images & 805 \\
      Editing Cases & 805 \\
      Sub-instructions & 1,610 \\
      Avg. Resolution & $1193 \times 831$ \\
      Max. Resolution & $1630 \times 1420$ \\
      \bottomrule
    \end{tabularx}
  \end{minipage}
  \hfill 
  \begin{minipage}[c]{0.6\textwidth} 
    \centering
    \includegraphics[width=\textwidth]{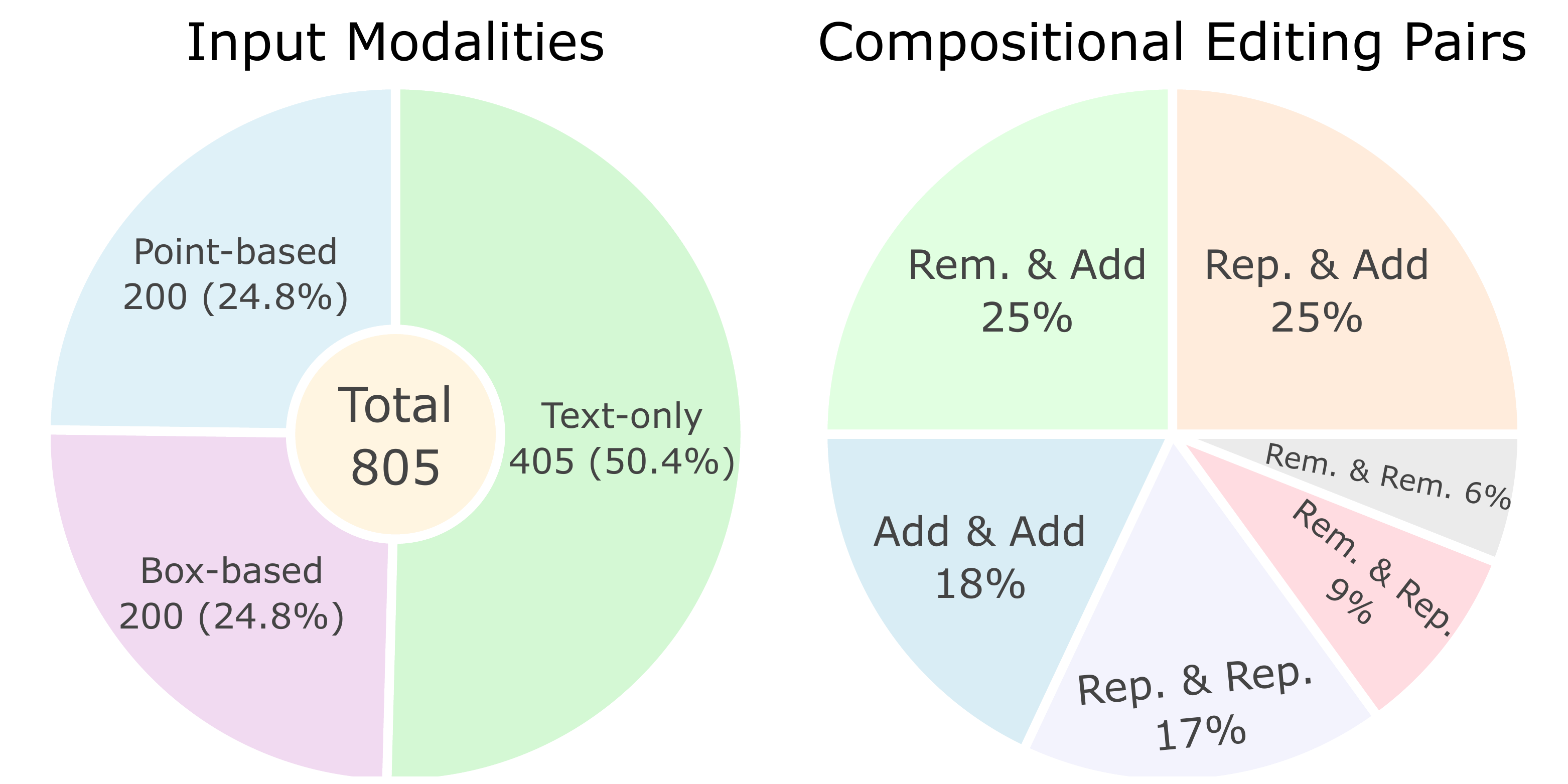}
  \end{minipage}

\caption{\textbf{Statistical overview of \ourbench{}}. It contains 805 cases and 1,610 sub-instructions (left), stratified across three input modalities (middle). The right chart shows the distribution of compositional editing pairs (Rem.: Remove, Rep.: Replace).}
\vspace{-3mm}
  \label{fig:combined_stat}
\end{figure}

We introduce \ourbench{} to evaluate compositional image editing, comprising 805 high-quality cases paired with sub-instructions. To assess robustness across grounding signals, the benchmark is stratified into point-based, box-based, and text-only modalities. Detailed statistics are presented in \cref{fig:combined_stat}.

\subsection{Data Collection}

\begin{algorithm}[t]
\caption{\textbf{Spatially Diverse Point Sampling}}
\label{alg:diverse_points}
\small
\begin{algorithmic}[1]
\Require Mask $M \in \{0,1\}^{H \times W}$, Count $K=4$
\Ensure Point set $\mathcal{S}$

\State $\mathcal{G} \gets \{ p \mid \forall q \in \mathcal{N}_p^{3\times3}, M(q)=1 \}$ \Comment{Interior pixels extraction}
\State $\mathbf{c} \gets \text{Mean}(\mathcal{G})$; Partition $\mathcal{G}$ into quadrants $\{Q_k\}_{k=1}^4$ centered at $\mathbf{c}$
\State $\mathcal{S} \gets \bigcup_{k=1}^4 \{ \arg\max_{p \in Q_k} \|p - \mathbf{c}\|_2 \text{ s.t. } Q_k \neq \emptyset \}$
\While{$|\mathcal{S}| < K$}
    \State $\mathcal{S} \gets \mathcal{S} \cup \{ \arg\max_{p \in \mathcal{G} \setminus \mathcal{S}} \sum_{s \in \mathcal{S}} \|p - s\|_2 \}$ \Comment{Maximize dispersion}
\EndWhile
\State \Return $\mathcal{S}$
\end{algorithmic}
\end{algorithm}

We construct \ourbench{} based on the OmniEdit dataset~\cite{wei2024omniedit}. From an initial pool, we filter for high-quality images and leverage GPT-4o to generate coherent compositional instructions involving combinations of replace, add, and remove operations. We strictly enforce order-invariance by ensuring that executing sub-instructions in any sequence yields the same result. We then conduct rigorous human verification to eliminate ambiguous samples, resulting in 805 refined pairs.

To generate spatial grounding signals, we randomly sample subsets for point and box annotations. For point-based cases, we employ a spatial diversity strategy (Algorithm~\ref{alg:diverse_points}) to select four distinct foreground points, ensuring coverage of the object's extent. For box-based cases, we compute the minimal bounding box enclosing the target mask. This process yields a diverse benchmark covering point, box, and text modalities to facilitate comprehensive evaluations.



\subsection{Evaluation Metrics}

To rectify generation-induced spatial shifts, we follow prior work~\cite{qian2025gie} and initially align the edited image to the original utilizing SIFT keypoints~\cite{lowe2004distinctive}, FLANN-based matching~\cite{muja2009fast}, and affine transformations. Within this aligned space, to enable fine-grained evaluation, we explicitly distinguish between edited and non-edited regions localized via our grounding module. Departing from static or coarse mask definitions common in prior work~\cite{jupnp, zhang2023magicbrush, qian2025gie}, we dynamically determine the edited region based on the operation type: it comprises the union of source and target subjects for \emph{Replace}, the target subject for \emph{Add}, and the source subject for \emph{Remove}, while the non-edited region acts as the spatial complement. To ensure a robust and comprehensive evaluation within these precisely localized regions, following ImgEdit~\cite{ye2025imgedit}, we independently leverage GPT-4o and Gemini-2.5-Pro~\cite{comanici2025gemini} to assess \emph{Semantic Correctness} (SC) and \emph{Perceptual Quality} (PQ). Importantly, we enforce PQ to be no greater than SC, reflecting the principle that visual quality is only meaningful when the editing instructions are semantically satisfied. Conversely, for non-edited regions, we evaluate background preservation using pixel-level metrics, including MSE, PSNR, and SSIM, to objectively measure the retention of original content throughout the editing process.

\section{Experiments}
\label{sec:experiments}

\begin{table}[t]
\centering
\caption{\textbf{Performance comparison on \ourbench{}}. Best results within each group are highlighted in \textbf{bold}. \colorbox{graybg}{Gray} indicates our method. Notably, our \ourmethod{} integrated with Lumina-DiMOO achieves state-of-the-art performance among training-free methods, steadily narrowing the performance gap to top-tier fully supervised models.}
\label{tab:main_result}
\small
\scalebox{0.94}{
\begin{tabular}{l c c c c c c c c c c}
\toprule
\multirow{2}{*}{\textbf{Method}} & &
\multicolumn{3}{c}{\textbf{Non-edit}} & \phantom{} &
\multicolumn{2}{c}{$\textbf{Edit}_{\mathrm{GPT}}$} & \phantom{} &
\multicolumn{2}{c}{$\textbf{Edit}_{\mathrm{Gemini}}$}\\
\cmidrule{3-5} \cmidrule{7-8} \cmidrule{10-11} 
& & \small \textbf{MSE $\downarrow$} & \small \textbf{ PSNR $\uparrow$} &  \small \textbf{ SSIM $\uparrow$} & & \small \textbf{  SC $\uparrow$} & \small \textbf{ PQ $\uparrow$} & & \small \textbf{  SC $\uparrow$} & \small \textbf{ PQ $\uparrow$} \\

\midrule
\multicolumn{11}{c}{\textit{End-to-End Image Editing Models}} \\
\midrule
OneDiffusion~\cite{le2025one} & & 1247.74 & 20.61 & 0.6831 &  & 1.81 & 1.79 & & 1.94 & 1.81\\
InstructPix2Pix~\cite{brooks2023instructpix2pix} & & 2540.06 & 17.71 & 0.7108 &  & 1.89 & 1.86 & & 1.87 & 1.68 \\
MagicBrush~\cite{zhang2023magicbrush} & & 2410.44 & 19.36 & 0.7210 &  & 2.13 & 2.04 & & 2.08 & 1.74 \\
Lumina-DiMOO~\cite{xin2025lumina} & & 1208.22 & 19.80 & 0.6461 &  & 3.10 & 2.92 & & 3.10 & 2.34\\
OmniGen2~\cite{wu2025omnigen2} & & 1927.87 & 19.88 & 0.7368 &  & 3.65 & 3.36 & & 3.67 & 3.13\\
FLUX.1-Kontext~\cite{labs2025flux} & & 2321.48 & 16.91 & 0.5900 &  & 3.94 & 3.61 & & 3.94 & 3.35 \\
Qwen-Image~\cite{wu2025qwen} & & 1758.97 & 21.20 & 0.7902 &  & 4.38 & 4.15 & & 4.33 & 3.98 \\
LongCat~\cite{team2025longcat} & & 1166.27 & 20.26 & 0.7266 &  & 4.51 & 4.33 & & 4.52 & 4.18 \\
Edit-R1~\cite{li2025uniworldv2} & & 1557.25 & 18.81 & 0.7083 &  & 4.64 & 4.46 & & 4.55 & 4.19\\
Nano-Banana-1~\cite{nano-banana-1} & & \textbf{687.79} & \textbf{23.83} & \textbf{0.8338} &  & 4.48 & 4.23 & & 4.56 & 4.35\\
GPT-Image-1~\cite{gptimage1} & & 5080.22 & 12.31 & 0.4714 &  & \textbf{4.71} & \textbf{4.66} & & \textbf{4.60} & \textbf{4.46}\\

\midrule
\multicolumn{11}{c}{\textit{Training-free Editing Methods applied to Base Models}} \\
\midrule
DirectInversion+P2P~\cite{jupnp} & & 3008.64 & 14.54 & 0.5848 &  & 2.04 & 2.00 & & 2.07 & 1.75\\
DirectInversion+PnP~\cite{jupnp} & & 2126.71 & 16.33 & 0.6404 &  & 2.19 & 2.15 & & 2.15 & 1.93 \\
DICE+Lumina-DiMOO~\cite{hedice} & & 8323.89 & 9.38 & 0.3866 &  & 2.81 & 2.71 & & 2.94 & 2.45\\
\rowcolor{graybg}
\ourmethod{}+MMaDA & & 3891.24 & 14.00 & 0.5522 & & 2.96 & 2.80 & & 2.74 & 2.54\\
\rowcolor{graybg}
\ourmethod{}+Lumina-DiMOO &  &\textbf{1224.89} & \textbf{20.40} & \textbf{0.7083} & & \textbf{4.47} & \textbf{3.98} & & \textbf{4.26} & \textbf{3.78} \\

\bottomrule
\end{tabular}
}
\end{table}

\subsection{Experimental Setup}
\label{sec:exp_setup}

As a general training-free image editing method, we build \ourmethod{} upon two popular DLLMs, Lumina-DiMOO~\cite{xin2025lumina} and MMaDA~\cite{yang2025mmada}, which are original developed for text-to-image generation. We compare \ourmethod{} with the following baselines:
\begin{itemize}
    \item \textbf{Training-free methods}, including DICE~\cite{hedice}\footnote{Since DICE does not open-source its code, we implement it by ourself based on a DLLM Lumina-DiMOO.}, Direct Inversion with Plug-and-Play (PnP)~\cite{tumanyan2023plug} and Prompt-to-Prompt (P2P)~\cite{hertzprompt}\footnote{Both PnP and P2P are based on Stable-Diffusion-v1.4~\cite{Rombach_2022_CVPR}.}.
    \item \textbf{Training-based methods}, including closed-source (GPT-Image-1~\cite{gptimage1}, Nano-Banana-1~\cite{nano-banana-1}) and open-source models (Edit-R1~\cite{li2025uniworldv2}, LongCat~\cite{team2025longcat}, Qwen-Image~\cite{wu2025qwen}, FLUX.1-Kontext~\cite{labs2025flux}, OmniGen2~\cite{wu2025omnigen2}, MagicBrush~\cite{zhang2023magicbrush}, InstructPix2Pix~\cite{brooks2023instructpix2pix}, and OneDiffusion~\cite{le2025one}), alongside Lumina-DiMOO's official pipeline as a strong baseline.
\end{itemize}
To align with their respective input formats, \ourmethod{} and training-free baselines adopt multi-instructions step-by-step, while end-to-end models process multi-instruction inputs jointly. Similarly, training-free methods use global descriptions extracted via GPT-4o~\cite{gpt4o}, and instruction-guided end-to-end models directly use raw instructions. Closed-source models are accessed via API calls, all other models or methods are evaluated on a single A100 GPU. Additionally, all evaluated models employ their respective default hyperparameters in our experiments.

\begin{figure*}[t]
  \centering
  \includegraphics[width=\linewidth]{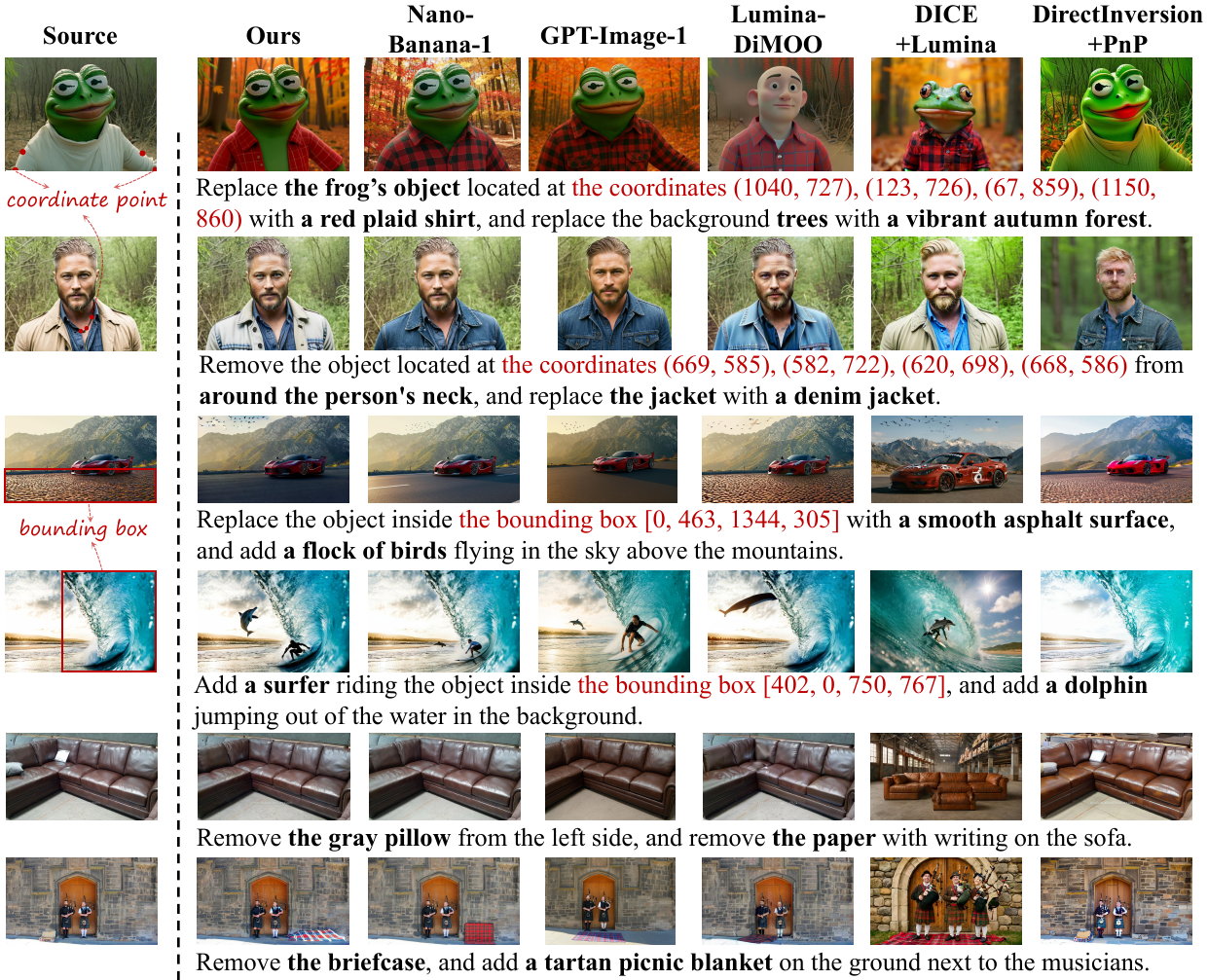}
\caption{\textbf{Qualitative comparison on \ourbench{}}. \ourmethod{} accurately follows instructions and preserves structural fidelity, avoiding unintended alterations (e.g., DICE) and aspect ratio distortions (e.g., GPT-Image-1) while achieving photorealistic consistency. Zoom in for better view.}
  \label{fig:qualitative_result}
  \vspace{-3mm}
\end{figure*}

\subsection{Experimental Results}

Table~\ref{tab:main_result} shows the quantitative evaluation of our proposed \ourmethod{} framework on \ourbench{}, compared against three representative training-free methods and eleven supervised end-to-end models to demonstrate its robust capabilities.

\noindent\textbf{Effectiveness of Editing on Target Regions.}
Our \ourmethod{} framework demonstrates superior performance in \emph{SC} while maintaining high \emph{PQ}\footnote{All SC and PQ scores reported herein are averaged between GPT and Gemini.}. We detail its advantages across several dimensions:
\begin{itemize}
    \item \textbf{Superiority over Training-Free Methods:} \ourmethod{} achieves remarkable gains over training-free baselines, surpassing DICE by 51.83\% in SC and 50.39\% in PQ. This margin extends further against other top-performing training-free methods (101.15\% in SC, 90.20\% in PQ), confirming that coupling discrete inversion with precise grounding is essential for robust editing.
    \item \textbf{Competitiveness with Supervised End-to-End Models:} Remarkably, \ourmethod{} surpasses the average performance of the 11 supervised models by 22.49\% in SC and 17.54\% in PQ. This demonstrates that a training-free framework can deliver highly competitive results, steadily closing the distance to top-performing closed-source models like Nano-Banana-1.
     \item \textbf{Addressing Base Model Limitations:} The image-to-image baseline of Lumina-DiMOO often misinterprets prompts and fails to accurately follow instructions (\cref{fig:qualitative_result}). By resolving these ambiguities through grounded inversion, \ourmethod{} significantly boosts SC by 40.81\% and PQ by 47.53\%.
    \item \textbf{Impact of Backbone Selection:} Weaker backbones like MMaDA struggle with fixed resolutions and complex operations (e.g., unnatural \emph{Add}/\emph{Remove} transitions). Adopting the powerful Lumina-DiMOO allows \ourmethod{} to overcome these bottlenecks, ensuring consistent, high-quality edits.
\end{itemize}

\begin{table}[htbp]
    \centering
    \caption{\textbf{Evaluation on ImgEdit~\cite{ye2025imgedit} benchmark}. Rep.: Replace, Rem.: Remove. }
    \label{tab:image_editing_result}
    \begin{minipage}[t]{0.48\textwidth}
        \centering
    
    \small
    \begin{tabular}{lccc}
    \toprule
    \textbf{Method} & \textbf{Rep.$\uparrow$} & \textbf{Add$\uparrow$} & \textbf{Rem.$\uparrow$} \\
    \midrule
    
    MagicBrush~\cite{zhang2023magicbrush} & 1.97 & 2.84 & 1.58 \\
    AnyEdit~\cite{yu2025anyedit} & 2.47 & 3.18 & 2.23 \\
    UltraEdit~\cite{zhao2024ultraedit} & 2.96 & 3.44 & 1.45 \\
    ICEdit~\cite{zhang2025enabling} & 3.15 & 3.58 & 2.93 \\
    Step1X-Edit~\cite{liu2025step1x} & \textbf{3.40} & \textbf{3.88} & \textbf{2.41} \\

    \bottomrule
    \end{tabular}
    \end{minipage}
    \hfill %
    \begin{minipage}[t]{0.48\textwidth}
    \small
    \begin{tabular}{lccc}
    \toprule
    \textbf{Method} & \textbf{Rep.$\uparrow$} & \textbf{Add$\uparrow$} & \textbf{Rem.$\uparrow$} \\
    \midrule

    OmniGen~\cite{xiao2025omnigen} & 2.94 & 3.47 & 2.43 \\
    BAGEL~\cite{deng2025emerging} & 3.30 & 3.56 & 2.62 \\
    UniWorld-V1~\cite{lin2025uniworld} & 3.47 & 3.82 & 3.24 \\
    Lumina~\cite{xin2025lumina} & 3.83 & 3.82 & 2.76 \\
    \rowcolor{graybg} \textbf{\ourmethod{} (Ours)} & \textbf{4.22} & \textbf{3.90} & \textbf{3.84} \\
    
    \bottomrule
    \end{tabular}
    \end{minipage}
\end{table}

\noindent\textbf{Preservation of Non-edited Regions.}
\label{5.2.2:Preservation of Non-edited Regions.}
Beyond editing accuracy, \ourmethod{} exhibits exceptional capability in preserving non-edited content, effectively avoiding the pitfalls of previous methods:
\begin{itemize}
    \item \textbf{Advantage over Global Inversion:} Methods like DICE treat logits from a single forward step as the ground-truth $\boldsymbol{y}_0$, perturbing the token distribution and severely degrading background fidelity. By leveraging precise localization, \ourmethod{} reduces MSE by 85.28\% against DICE, while improving PSNR by 117.48\% and SSIM by 83.21\%, ensuring superior visual consistency.
    \item \textbf{Optimal Performance Trade-off:} Compared to models heavily optimized for editing (such as GPT-Image-1), \ourmethod{} offers a much better balance. It prevents background distortion by reducing MSE by 75.89\% and boosting PSNR and SSIM by 65.72\% and 50.25\%, respectively.
    \item \textbf{Fidelity Constraints of DLLMs:} While \ourmethod{} effectively preserves visual semantics, its low-level fidelity still lags behind Nano-Banana-1. This is attributed to the inherent reconstruction loss of the VQModel in Lumina-DiMOO. As DLLM autoencoders evolve, this gap will naturally close.
\end{itemize}

\subsection{Qualitative Comparison}

Figure~\ref{fig:qualitative_result} compares \ourmethod{} against baselines across six combinations of \emph{Replace}, \emph{Add}, and \emph{Remove} tasks. \ourmethod{} accurately executes editing instructions while strictly preserving unedited regions and structural fidelity. In contrast, training-free methods (e.g., DICE) and Lumina-DiMOO lack precise grounding, which leads to prominent artifacts, whereas GPT-Image-1 alters original aspect ratios and introduces structural distortions. Furthermore, \ourmethod{} ensures exceptional photorealism and physical consistency by leveraging its discrete inversion to inject source priors (e.g., lighting and texture). For instance, as shown in Row 1, \ourmethod{} retains the original shallow depth of field, avoiding the unnatural background sharpening observed in Nano-Banana-1. Similarly, in Row 4, it accurately renders a backlit silhouette for the inserted surfer, effectively mitigating the discordant ``copy-paste'' artifacts produced by Nano-Banana-1 and GPT-Image-1.

\subsection{Evaluation on ImgEdit Benchmark}

To validate the broad applicability of our framework, we further evaluate \ourmethod{} on the ImgEdit benchmark across three core tasks: \emph{Replace}, \emph{Add}, and \emph{Remove}. The evaluation metrics are computed using GPT-4o. We compare our approach against Lumina-DiMOO's image-to-image pipeline, alongside other representative methods (MagicBrush, AnyEdit~\cite{yu2025anyedit}, UltraEdit~\cite{zhao2024ultraedit}, ICEdit~\cite{zhang2025enabling}, Step1X-Edit~\cite{liu2025step1x}, OmniGen~\cite{xiao2025omnigen}, BAGEL~\cite{deng2025emerging}, and UniWorld-V1~\cite{lin2025uniworld}). As shown in Table~\ref{tab:image_editing_result}, \ourmethod{} consistently outperforms the Lumina-DiMOO baseline, achieving relative improvements of 10.18\%, 2.09\%, and 39.13\% on the respective tasks. Notably, the substantial gain in the \emph{Remove} task highlights the efficacy of our precise grounding and specialized inversion strategy, effectively overcoming the baseline's limitations in cleanly erasing objects. By outperforming all evaluated baselines, \ourmethod{} demonstrates robust performance and consistent effectiveness.

\begin{figure}[tb]
  \centering
  \includegraphics[width=\linewidth]{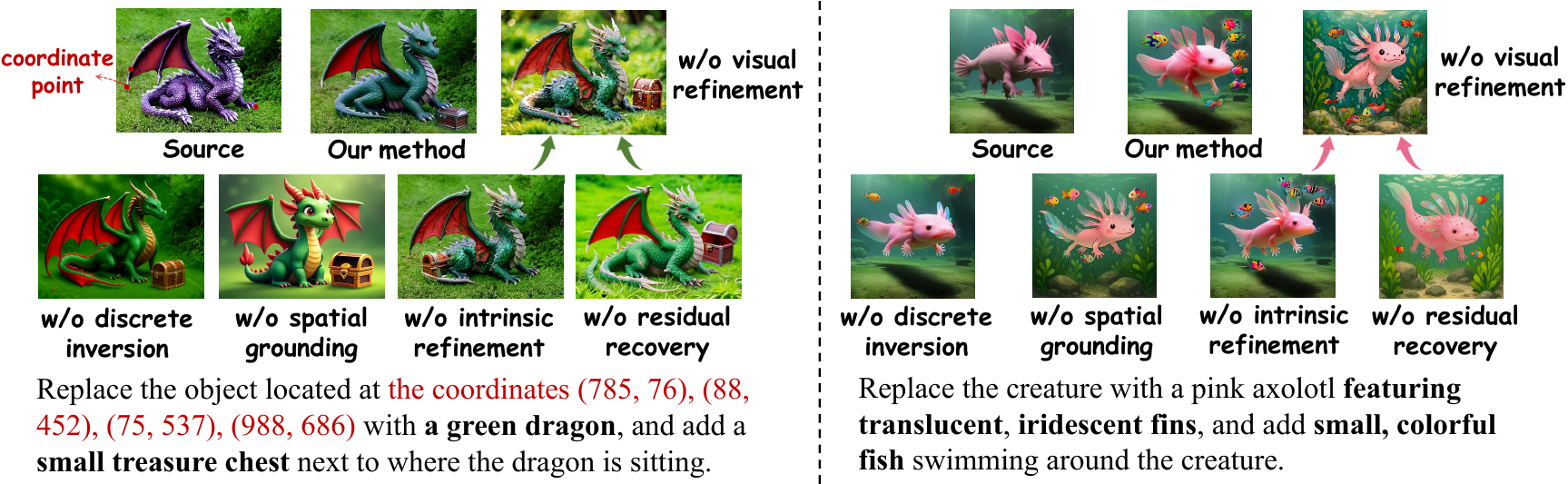}
    \caption{\textbf{Visual ablation on the proposed components}. Removing the \textbf{discrete inversion} degrades structural fidelity and distorts object shapes. Omitting the \textbf{spatial grounding} leads to the loss of background context. Disabling the \textbf{visual refinement} blurs local details and compromises the seamless preservation of the unedited regions.}
  \label{fig:ablation}
\end{figure}

\begin{table}[t]
\centering
\caption{\textbf{Ablation experiments}. We compare the full method against variants with specific modules removed. The results confirm that the full framework achieves the best performance in both background preservation and editing quality.}
\label{tab:ablation_study}

\scalebox{0.86}{
\begin{tabular}{l lllllllll}
\toprule
\multirow{2}{*}{\textbf{Method}} &
\multicolumn{3}{c}{\textbf{Non-edit}} & \phantom{} &
\multicolumn{2}{c}{$\textbf{Edit}_{\mathrm{GPT}}$} & \phantom{} &
\multicolumn{2}{c}{$\textbf{Edit}_{\mathrm{Gemini}}$}\\

\cmidrule{2-4} \cmidrule{6-7} \cmidrule{9-10} 
& \small \textbf{MSE $\downarrow$} & \small \textbf{ PSNR $\uparrow$} &  \small \textbf{ SSIM $\uparrow$} & & \small \textbf{  SC $\uparrow$} & \small \textbf{ PQ $\uparrow$} & & \small \textbf{  SC $\uparrow$} & \small \textbf{ PQ $\uparrow$} \\
\midrule
\rowcolor{graybg}
\textbf{Ours (Full)} & \textbf{1224.89} & \textbf{20.40} & \textbf{0.7083} & & \textbf{4.47} & \textbf{3.98} & & \textbf{4.26} & \textbf{3.78} \\

w/o discrete inversion & 2196.29$_{\textcolor{lightred}{79\%\uparrow}}$ & 17.85$_{\textcolor{lightred}{13\%\downarrow}}$ & 0.6628$_{\textcolor{lightred}{6\%\downarrow}}$ & & 4.30$_{\textcolor{lightred}{4\%\downarrow}}$ & 3.83$_{\textcolor{lightred}{4\%\downarrow}}$ & & 4.08$_{\textcolor{lightred}{4\%\downarrow}}$ & 3.28$_{\textcolor{lightred}{13\%\downarrow}}$ \\    

w/o spatial grounding & 8446.17$_{\textcolor{lightred}{590\%\uparrow}}$ & 9.42$_{\textcolor{lightred}{54\%\downarrow}}$ & 0.4161$_{\textcolor{lightred}{41\%\downarrow}}$ & & 3.60$_{\textcolor{lightred}{19\%\downarrow}}$ & 3.43$_{\textcolor{lightred}{14\%\downarrow}}$ & & 2.68$_{\textcolor{lightred}{37\%\downarrow}}$ & 2.21$_{\textcolor{lightred}{42\%\downarrow}}$ \\

w/o visual refinement & 2476.78$_{\textcolor{lightred}{102\%\uparrow}}$ & 17.23$_{\textcolor{lightred}{16\%\downarrow}}$ & 0.6444$_{\textcolor{lightred}{9\%\downarrow}}$ & & 3.73$_{\textcolor{lightred}{17\%\downarrow}}$ & 3.25$_{\textcolor{lightred}{18\%\downarrow}}$ & & 3.56$_{\textcolor{lightred}{16\%\downarrow}}$ & 2.67$_{\textcolor{lightred}{29\%\downarrow}}$ \\

w/o intrinsic refinement & 1424.29$_{\textcolor{lightred}{16\%\uparrow}}$ & 19.80$_{\textcolor{lightred}{3\%\downarrow}}$ & 0.6987$_{\textcolor{lightred}{1\%\downarrow}}$ & & 3.96$_{\textcolor{lightred}{11\%\downarrow}}$ & 3.51$_{\textcolor{lightred}{12\%\downarrow}}$ & & 3.87$_{\textcolor{lightred}{9\%\downarrow}}$ & 3.11$_{\textcolor{lightred}{18\%\downarrow}}$\\

w/o residual recovery & 2316.94$_{\textcolor{lightred}{89\%\uparrow}}$ & 17.65$_{\textcolor{lightred}{13\%\downarrow}}$ & 0.6606$_{\textcolor{lightred}{7\%\downarrow}}$ & & 4.16$_{\textcolor{lightred}{7\%\downarrow}}$ & 3.80$_{\textcolor{lightred}{5\%\downarrow}}$ & & 3.85$_{\textcolor{lightred}{10\%\downarrow}}$ & 3.21$_{\textcolor{lightred}{15\%\downarrow}}$ \\

\bottomrule
\end{tabular}}
\vspace{-3mm}
\end{table}


\subsection{Ablation Study}

We evaluate our proposed components through an ablation study (Fig.~\ref{fig:ablation}, Table~\ref{tab:ablation_study}). First, the \textbf{discrete inversion} is critical for structural consistency. Substituting it with standard inpainting degrades semantic metrics (SC/PQ) and compromises fine-grained geometry, visibly distorting the dragon and thinning the axolotl. Second, the \textbf{spatial grounding} is fundamental for background preservation; its absence forces global editing, causing a 589.55\% MSE increase and severe background hallucination (e.g., the grassland). Finally, the \textbf{visual refinement} ensures high-fidelity results and seamless integration. Within it, omitting \emph{intrinsic refinement} blurs textures, while removing \emph{residual recovery} causes distinct boundary artifacts. Disabling this entire module incurs a 102.20\% MSE penalty and a 23.71\% PQ decline, confirming its necessity for realistic editing.

\subsection{Sensitivity Analysis}

\begin{table}[t]
\centering
\caption{\textbf{Sensitivity analysis} of the mixing coefficient $\lambda$. Best scores are highlighted in \textbf{bold}. The default setting $\lambda = 0.2$ is highlighted in gray. \ourmethod{} achieves optimal editing quality at $\lambda = 0.2$ with stable performance across the range.}
\label{tab:sensitivity_analysis}

\begin{tabular}{l c c c c c c c c c c}
\toprule
\multirow{2}{*}{\textbf{$\lambda$}} & \phantom{} &
\multicolumn{3}{c}{\textbf{Non-edit}} & \phantom{} &
\multicolumn{2}{c}{$\textbf{Edit}_{GPT}$} & \phantom{} &
\multicolumn{2}{c}{$\textbf{Edit}_{Gemini}$}\\
\cmidrule{3-5} \cmidrule{7-8} \cmidrule{10-11} 
 & & \small \textbf{MSE $\downarrow$} & \small \textbf{ PSNR $\uparrow$} &  \small \textbf{ SSIM $\uparrow$} & & \small \textbf{  SC $\uparrow$} & \small \textbf{ PQ $\uparrow$} & & \small \textbf{  SC $\uparrow$} & \small \textbf{ PQ $\uparrow$} \\
\midrule
0.0 & & \textbf{1224.62} & 20.35 & \textbf{0.7086} & & 4.30 & 3.86 & & 4.12 & 3.71\\
0.1 & & 1306.47 & 20.22 & 0.7066 & & 4.26 & 3.82 & & 4.12 & 3.69 \\    
\rowcolor{graybg}
0.2 & & 1224.89 & 20.40 & 0.7083 &  & \textbf{4.47} & \textbf{3.98} & & \textbf{4.26} & \textbf{3.78} \\
0.3 & & 1308.57 & 20.33 & 0.7075 &  & 4.28 & 3.85 & & 4.09 & 3.70\\
0.4 & & 1225.40 & \textbf{20.51} & 0.7084 &  & 4.21 & 3.83 & & 4.03 & 3.75 \\
\bottomrule
\end{tabular}
\end{table}

To investigate the impact of the inversion residual strength on editing quality, we analyze the sensitivity of the mixing coefficient $\lambda$. As summarized in Table~\ref{tab:sensitivity_analysis}, setting $\lambda = 0.2$ yields the best performance on edited regions while maintaining near-optimal metrics in non-edited areas, striking an effective balance between semantic editing and content preservation. Furthermore, performance remains consistently stable across the range of $\lambda \in [0, 0.4]$ with only marginal fluctuations, demonstrating the robustness of \ourmethod{} to parameter variations.



\section{Conclusions and Limitations}

In this work, we presented \ourmethod{}, the first training-free image editing framework specifically tailored for DLLMs. By introducing a principled discrete inversion mechanism alongside a compositional grounding and refinement pipeline, \ourmethod{} effectively bridges the gap between discrete representation and high-fidelity image editing. To rigorously evaluate these capabilities, we established \ourbench{}, a comprehensive benchmark emphasizing compositional instructions and diverse grounding modalities. Extensive experiments demonstrate that \ourmethod{} significantly outperforms existing training-free baselines and achieves photorealistic consistency comparable to state-of-the-art supervised methods.

However, the current performance is bounded by the precision of off-the-shelf segmentation models and the inherent reconstruction loss of VQ-based tokenizers. Nevertheless, designed as a generic framework, \ourmethod{} is expected to naturally overcome these bottlenecks as foundational segmentation models and DLLM architectures continue to evolve, paving the way for high-fidelity unified multi-modal editing in future research endeavors.


%
%
\bibliographystyle{splncs04}
\bibliography{main}

@String(CVPR  = {IEEE Conf. Comput. Vis. Pattern Recog.})

@String(CVPR  = {CVPR})

@inproceedings{teweladd,
  title={Add-it: Training-Free Object Insertion in Images With Pretrained Diffusion Models},
  author={Tewel, Yoad and Gal, Rinon and Samuel, Dvir and Atzmon, Yuval and Wolf, Lior and Chechik, Gal},
  booktitle={The Thirteenth International Conference on Learning Representations},
year={2025}
}

@article{dhariwal2021diffusion,
  title={Diffusion models beat gans on image synthesis},
  author={Dhariwal, Prafulla and Nichol, Alexander},
  journal={Advances in neural information processing systems},
  volume={34},
  pages={8780--8794},
  year={2021}
}

@inproceedings{songdenoising,
  title={Denoising Diffusion Implicit Models},
  author={Song, Jiaming and Meng, Chenlin and Ermon, Stefano},
  booktitle={International Conference on Learning Representations},
year={2021}
}

@inproceedings{mokady2023null,
  title={Null-text inversion for editing real images using guided diffusion models},
  author={Mokady, Ron and Hertz, Amir and Aberman, Kfir and Pritch, Yael and Cohen-Or, Daniel},
  booktitle={Proceedings of the IEEE/CVF conference on computer vision and pattern recognition},
  pages={6038--6047},
  year={2023}
}

@inproceedings{jupnp,
  title={PnP Inversion: Boosting Diffusion-based Editing with 3 Lines of Code},
  author={Ju, Xuan and Zeng, Ailing and Bian, Yuxuan and Liu, Shaoteng and Xu, Qiang},
  booktitle={The Twelfth International Conference on Learning Representations},
    year={2024}
}

@article{dao2025discrete,
  title={Discrete Noise Inversion for Next-scale Autoregressive Text-based Image Editing},
  author={Dao, Quan and He, Xiaoxiao and Han, Ligong and Nguyen, Ngan Hoai and Nobar, Amin Heyrani and Ahmed, Faez and Zhang, Han and Nguyen, Viet Anh and Metaxas, Dimitris},
  journal={arXiv preprint arXiv:2509.01984},
  year={2025}
}

@inproceedings{wei2024omniedit,
  title={Omniedit: Building image editing generalist models through specialist supervision},
  author={Wei, Cong and Xiong, Zheyang and Ren, Weiming and Du, Xeron and Zhang, Ge and Chen, Wenhu},
  booktitle={The Thirteenth International Conference on Learning Representations},
  year={2025}
}

@article{ye2025imgedit,
  title={Imgedit: A unified image editing dataset and benchmark},
  author={Ye, Yang and He, Xianyi and Li, Zongjian and Lin, Bin and Yuan, Shenghai and Yan, Zhiyuan and Hou, Bohan and Yuan, Li},
  journal={arXiv preprint arXiv:2505.20275},
  year={2025}
}

@misc{gpt4o,
author={OpenAI},
title={Hello GPT-4o},
year={2024},
howpublished={News announcement by OpenAI},
url={https://openai.com/index/hello-gpt-4o/}
}

@article{yang2025mmada,
  title={Mmada: Multimodal large diffusion language models},
  author={Yang, Ling and Tian, Ye and Li, Bowen and Zhang, Xinchen and Shen, Ke and Tong, Yunhai and Wang, Mengdi},
  journal={arXiv preprint arXiv:2505.15809},
  year={2025}
}

@article{hedice,
  title={DICE: Discrete Inversion Enabling Controllable Editing for Masked Generative Models},
  author={He, Xiaoxiao and Han, Ligong and Quan Dao, Song Wen and Bai, Minhao and Di Liu, Han Zhang and Juefei-Xu, Felix and Tan, Chaowei and Liu, Bo and Min, Martin Renqiang and Li, Kang and others},
year={2025}
}

@inproceedings{tumanyan2023plug,
  title={Plug-and-play diffusion features for text-driven image-to-image translation},
  author={Tumanyan, Narek and Geyer, Michal and Bagon, Shai and Dekel, Tali},
  booktitle={Proceedings of the IEEE/CVF conference on computer vision and pattern recognition},
  pages={1921--1930},
  year={2023}
}

@inproceedings{hertzprompt,
  title={Prompt-to-Prompt Image Editing with Cross-Attention Control},
  author={Hertz, Amir and Mokady, Ron and Tenenbaum, Jay and Aberman, Kfir and Pritch, Yael and Cohen-or, Daniel},
  booktitle={The Eleventh International Conference on Learning Representations},
  year={2023}
}

@InProceedings{Rombach_2022_CVPR,
    author    = {Rombach, Robin and Blattmann, Andreas and Lorenz, Dominik and Esser, Patrick and Ommer, Bj\"orn},
    title     = {High-Resolution Image Synthesis With Latent Diffusion Models},
    booktitle = {Proceedings of the IEEE/CVF Conference on Computer Vision and Pattern Recognition (CVPR)},
    month     = {June},
    year      = {2022},
    pages     = {10684-10695}
}

@misc{gptimage1,
  author       = {OpenAI},
  title        = {GPT Image 1: State-of-the-art image generation model},
  howpublished = {\url{https://platform.openai.com/docs/models/gpt-image-1}},
  year         = {2025}
}

@misc{nano-banana-1,
  title        = {Gemini 2.5 Flash Image (Nano Banana)},
  author       = {Google},
  year         = {2025},
  howpublished = {\url{https://aistudio.google.com/models/gemini-2-5-flash-image}},
}

@article{li2025uniworldv2,
    title={Uniworld-V2: Reinforce Image Editing with Diffusion Negative-aware Finetuning and MLLM Implicit Feedback},
    author={Li, Zongjian and Liu, Zheyuan and Zhang, Qihui and Lin, Bin and Yuan, Shenghai and Yan, Zhiyuan and Ye, Yang and Yu, Wangbo and Niu, Yuwei and Yuan, Li},
    journal={arXiv preprint arXiv:2510.16888},
    year={2025}
}

@article{team2025longcat,
  title={Longcat-image technical report},
  author={Team, Meituan LongCat and Ma, Hanghang and Tan, Haoxian and Huang, Jiale and Wu, Junqiang and He, Jun-Yan and Gao, Lishuai and Xiao, Songlin and Wei, Xiaoming and Ma, Xiaoqi and others},
  journal={arXiv preprint arXiv:2512.07584},
  year={2025}
}

@article{wu2025qwen,
  title={Qwen-image technical report},
  author={Wu, Chenfei and Li, Jiahao and Zhou, Jingren and Lin, Junyang and Gao, Kaiyuan and Yan, Kun and Yin, Sheng-ming and Bai, Shuai and Xu, Xiao and Chen, Yilei and others},
  journal={arXiv preprint arXiv:2508.02324},
  year={2025}
}

@article{labs2025flux,
  title={FLUX. 1 Kontext: Flow Matching for In-Context Image Generation and Editing in Latent Space},
  author={Labs, Black Forest and Batifol, Stephen and Blattmann, Andreas and Boesel, Frederic and Consul, Saksham and Diagne, Cyril and Dockhorn, Tim and English, Jack and English, Zion and Esser, Patrick and others},
  journal={arXiv preprint arXiv:2506.15742},
  year={2025}
}

@article{wu2025omnigen2,
  title={OmniGen2: Exploration to Advanced Multimodal Generation},
  author={Wu, Chenyuan and Zheng, Pengfei and Yan, Ruiran and Xiao, Shitao and Luo, Xin and Wang, Yueze and Li, Wanli and Jiang, Xiyan and Liu, Yexin and Zhou, Junjie and others},
  journal={arXiv preprint arXiv:2506.18871},
  year={2025}
}

@article{zhang2023magicbrush,
  title={Magicbrush: A manually annotated dataset for instruction-guided image editing},
  author={Zhang, Kai and Mo, Lingbo and Chen, Wenhu and Sun, Huan and Su, Yu},
  journal={Advances in Neural Information Processing Systems},
  volume={36},
  pages={31428--31449},
  year={2023}
}

@inproceedings{brooks2023instructpix2pix,
  title={Instructpix2pix: Learning to follow image editing instructions},
  author={Brooks, Tim and Holynski, Aleksander and Efros, Alexei A},
  booktitle={Proceedings of the IEEE/CVF conference on computer vision and pattern recognition},
  pages={18392--18402},
  year={2023}
}

@inproceedings{le2025one,
  title={One diffusion to generate them all},
  author={Le, Duong H and Pham, Tuan and Lee, Sangho and Clark, Christopher and Kembhavi, Aniruddha and Mandt, Stephan and Krishna, Ranjay and Lu, Jiasen},
  booktitle={Proceedings of the Computer Vision and Pattern Recognition Conference},
  pages={2671--2682},
  year={2025}
}

@article{xin2025lumina,
  title={Lumina-dimoo: An omni diffusion large language model for multi-modal generation and understanding},
  author={Xin, Yi and Qin, Qi and Luo, Siqi and Zhu, Kaiwen and Yan, Juncheng and Tai, Yan and Lei, Jiayi and Cao, Yuewen and Wang, Keqi and Wang, Yibin and others},
  journal={arXiv preprint arXiv:2510.06308},
  year={2025}
}

@inproceedings{yu2025anyedit,
  title={Anyedit: Mastering unified high-quality image editing for any idea},
  author={Yu, Qifan and Chow, Wei and Yue, Zhongqi and Pan, Kaihang and Wu, Yang and Wan, Xiaoyang and Li, Juncheng and Tang, Siliang and Zhang, Hanwang and Zhuang, Yueting},
  booktitle={Proceedings of the Computer Vision and Pattern Recognition Conference},
  pages={26125--26135},
  year={2025}
}

@inproceedings{xiao2025omnigen,
  title={Omnigen: Unified image generation},
  author={Xiao, Shitao and Wang, Yueze and Zhou, Junjie and Yuan, Huaying and Xing, Xingrun and Yan, Ruiran and Li, Chaofan and Wang, Shuting and Huang, Tiejun and Liu, Zheng},
  booktitle={Proceedings of the Computer Vision and Pattern Recognition Conference},
  pages={13294--13304},
  year={2025}
}

@article{deng2025emerging,
  title={Emerging properties in unified multimodal pretraining},
  author={Deng, Chaorui and Zhu, Deyao and Li, Kunchang and Gou, Chenhui and Li, Feng and Wang, Zeyu and Zhong, Shu and Yu, Weihao and Nie, Xiaonan and Song, Ziang and others},
  journal={arXiv preprint arXiv:2505.14683},
  year={2025}
}

@article{lin2025uniworld,
  title={Uniworld: High-resolution semantic encoders for unified visual understanding and generation},
  author={Lin, Bin and Li, Zongjian and Cheng, Xinhua and Niu, Yuwei and Ye, Yang and He, Xianyi and Yuan, Shenghai and Yu, Wangbo and Wang, Shaodong and Ge, Yunyang and others},
  journal={arXiv preprint arXiv:2506.03147},
  year={2025}
}

@inproceedings{rombach2022high,
  title={High-resolution image synthesis with latent diffusion models},
  author={Rombach, Robin and Blattmann, Andreas and Lorenz, Dominik and Esser, Patrick and Ommer, Bj{\"o}rn},
  booktitle={Proceedings of the IEEE/CVF conference on computer vision and pattern recognition},
  pages={10684--10695},
  year={2022}
}

@inproceedings{cao2023masactrl,
  title={Masactrl: Tuning-free mutual self-attention control for consistent image synthesis and editing},
  author={Cao, Mingdeng and Wang, Xintao and Qi, Zhongang and Shan, Ying and Qie, Xiaohu and Zheng, Yinqiang},
  booktitle={Proceedings of the IEEE/CVF international conference on computer vision},
  pages={22560--22570},
  year={2023}
}

@article{tian2024visual,
  title={Visual autoregressive modeling: Scalable image generation via next-scale prediction},
  author={Tian, Keyu and Jiang, Yi and Yuan, Zehuan and Peng, Bingyue and Wang, Liwei},
  journal={Advances in neural information processing systems},
  volume={37},
  pages={84839--84865},
  year={2024}
}

@article{krojer2024learning,
  title={Learning action and reasoning-centric image editing from videos and simulation},
  author={Krojer, Benno and Vattikonda, Dheeraj and Lara, Luis and Jampani, Varun and Portelance, Eva and Pal, Chris and Reddy, Siva},
  journal={Advances in Neural Information Processing Systems},
  volume={37},
  pages={38035--38078},
  year={2024}
}

@article{liu2025step1x,
  title={Step1x-edit: A practical framework for general image editing},
  author={Liu, Shiyu and Han, Yucheng and Xing, Peng and Yin, Fukun and Wang, Rui and Cheng, Wei and Liao, Jiaqi and Wang, Yingming and Fu, Honghao and Han, Chunrui and others},
  journal={arXiv preprint arXiv:2504.17761},
  year={2025}
}

@article{qian2025gie,
  title={GIE-Bench: Towards Grounded Evaluation for Text-Guided Image Editing},
  author={Qian, Yusu and Lu, Jiasen and Fu, Tsu-Jui and Wang, Xinze and Chen, Chen and Yang, Yinfei and Hu, Wenze and Gan, Zhe},
  journal={arXiv preprint arXiv:2505.11493},
  year={2025}
}

@inproceedings{hessel2021clipscore,
  title={Clipscore: A reference-free evaluation metric for image captioning},
  author={Hessel, Jack and Holtzman, Ari and Forbes, Maxwell and Le Bras, Ronan and Choi, Yejin},
  booktitle={Proceedings of the 2021 conference on empirical methods in natural language processing},
  pages={7514--7528},
  year={2021}
}

@article{chen2025janus,
  title={Janus-pro: Unified multimodal understanding and generation with data and model scaling},
  author={Chen, Xiaokang and Wu, Zhiyu and Liu, Xingchao and Pan, Zizheng and Liu, Wen and Xie, Zhenda and Yu, Xingkai and Ruan, Chong},
  journal={arXiv preprint arXiv:2501.17811},
  year={2025}
}

@article{carion2025sam,
  title={Sam 3: Segment anything with concepts},
  author={Carion, Nicolas and Gustafson, Laura and Hu, Yuan-Ting and Debnath, Shoubhik and Hu, Ronghang and Suris, Didac and Ryali, Chaitanya and Alwala, Kalyan Vasudev and Khedr, Haitham and Huang, Andrew and others},
  journal={arXiv preprint arXiv:2511.16719},
  year={2025}
}

@inproceedings{ravisam,
  title={SAM 2: Segment Anything in Images and Videos},
  author={Ravi, Nikhila and Gabeur, Valentin and Hu, Yuan-Ting and Hu, Ronghang and Ryali, Chaitanya and Ma, Tengyu and Khedr, Haitham and R{\"a}dle, Roman and Rolland, Chloe and Gustafson, Laura and others},
  booktitle={The Thirteenth International Conference on Learning Representations},
  year={2025}
}

@article{xin2025resurrect,
  title={Resurrect mask autoregressive modeling for efficient and scalable image generation},
  author={Xin, Yi and Zhuo, Le and Qin, Qi and Luo, Siqi and Cao, Yuewen and Fu, Bin and He, Yangfan and Li, Hongsheng and Zhai, Guangtao and Liu, Xiaohong and others},
  journal={arXiv preprint arXiv:2507.13032},
  year={2025}
}

@article{liu2024lumina,
  title={Lumina-mgpt: Illuminate flexible photorealistic text-to-image generation with multimodal generative pretraining},
  author={Liu, Dongyang and Zhao, Shitian and Zhuo, Le and Lin, Weifeng and Xin, Yi and Li, Xinyue and Qin, Qi and Qiao, Yu and Li, Hongsheng and Gao, Peng},
  journal={arXiv preprint arXiv:2408.02657},
  year={2024}
}

@article{ge2024seed,
  title={Seed-data-edit technical report: A hybrid dataset for instructional image editing},
  author={Ge, Yuying and Zhao, Sijie and Li, Chen and Ge, Yixiao and Shan, Ying},
  journal={arXiv preprint arXiv:2405.04007},
  year={2024}
}

@article{zhao2024ultraedit,
  title={Ultraedit: Instruction-based fine-grained image editing at scale},
  author={Zhao, Haozhe and Ma, Xiaojian Shawn and Chen, Liang and Si, Shuzheng and Wu, Rujie and An, Kaikai and Yu, Peiyu and Zhang, Minjia and Li, Qing and Chang, Baobao},
  journal={Advances in Neural Information Processing Systems},
  volume={37},
  pages={3058--3093},
  year={2024}
}

@inproceedings{sheynin2024emu,
  title={Emu edit: Precise image editing via recognition and generation tasks},
  author={Sheynin, Shelly and Polyak, Adam and Singer, Uriel and Kirstain, Yuval and Zohar, Amit and Ashual, Oron and Parikh, Devi and Taigman, Yaniv},
  booktitle={Proceedings of the IEEE/CVF Conference on Computer Vision and Pattern Recognition},
  pages={8871--8879},
  year={2024}
}

@inproceedings{nguyen2025h,
  title={h-Edit: Effective and Flexible Diffusion-Based Editing via Doob's h-Transform},
  author={Nguyen, Toan and Do, Kien and Kieu, Duc and Nguyen, Thin},
  booktitle={Proceedings of the Computer Vision and Pattern Recognition Conference},
  pages={28490--28501},
  year={2025}
}

@inproceedings{fu2025feededit,
  title={Feededit: Text-based image editing with dynamic feedback regulation},
  author={Fu, Fengyi and Zhang, Lei and Huang, Mengqi and Mao, Zhendong},
  booktitle={Proceedings of the Computer Vision and Pattern Recognition Conference},
  pages={2661--2670},
  year={2025}
}

@inproceedings{chung2024style,
  title={Style injection in diffusion: A training-free approach for adapting large-scale diffusion models for style transfer},
  author={Chung, Jiwoo and Hyun, Sangeek and Heo, Jae-Pil},
  booktitle={Proceedings of the IEEE/CVF conference on computer vision and pattern recognition},
  pages={8795--8805},
  year={2024}
}

@inproceedings{morita2025tkg,
  title={TKG-DM: Training-free Chroma Key Content Generation Diffusion Model},
  author={Morita, Ryugo and Frolov, Stanislav and Moser, Brian Bernhard and Shirakawa, Takahiro and Watanabe, Ko and Dengel, Andreas and Zhou, Jinjia},
  booktitle={Proceedings of the Computer Vision and Pattern Recognition Conference},
  pages={13031--13040},
  year={2025}
}

@inproceedings{avrahami2025stable,
  title={Stable flow: Vital layers for training-free image editing},
  author={Avrahami, Omri and Patashnik, Or and Fried, Ohad and Nemchinov, Egor and Aberman, Kfir and Lischinski, Dani and Cohen-Or, Daniel},
  booktitle={Proceedings of the Computer Vision and Pattern Recognition Conference},
  pages={7877--7888},
  year={2025}
}

@inproceedings{xu2025stylessp,
  title={Stylessp: Sampling startpoint enhancement for training-free diffusion-based method for style transfer},
  author={Xu, Ruojun and Xi, Weijie and Wang, XiaoDi and Mao, Yongbo and Cheng, Zach},
  booktitle={Proceedings of the Computer Vision and Pattern Recognition Conference},
  pages={18260--18269},
  year={2025}
}

@inproceedings{mo2024freecontrol,
  title={Freecontrol: Training-free spatial control of any text-to-image diffusion model with any condition},
  author={Mo, Sicheng and Mu, Fangzhou and Lin, Kuan Heng and Liu, Yanli and Guan, Bochen and Li, Yin and Zhou, Bolei},
  booktitle={Proceedings of the IEEE/CVF conference on computer vision and pattern recognition},
  pages={7465--7475},
  year={2024}
}

@inproceedings{zhu2025training,
  title={Training-free Geometric Image Editing on Diffusion Models},
  author={Zhu, Hanshen and Zhu, Zhen and Zhang, Kaile and Gong, Yiming and Liu, Yuliang and Bai, Xiang},
  booktitle={Proceedings of the IEEE/CVF International Conference on Computer Vision},
  pages={19130--19140},
  year={2025}
}

@inproceedings{kim2025reflex,
  title={Reflex: Text-guided editing of real images in rectified flow via mid-step feature extraction and attention adaptation},
  author={Kim, Jimyeong and Park, Jungwon and Song, Yeji and Kwak, Nojun and Rhee, Wonjong},
  booktitle={Proceedings of the IEEE/CVF International Conference on Computer Vision},
  pages={15939--15948},
  year={2025}
}

@inproceedings{zhu2025kv,
  title={Kv-edit: Training-free image editing for precise background preservation},
  author={Zhu, Tianrui and Zhang, Shiyi and Shao, Jiawei and Tang, Yansong},
  booktitle={Proceedings of the IEEE/CVF International Conference on Computer Vision},
  pages={16607--16617},
  year={2025}
}

@inproceedings{hu2025anchor,
  title={Anchor token matching: Implicit structure locking for training-free ar image editing},
  author={Hu, Taihang and Li, Linxuan and Wang, Kai and Wang, Yaxing and Yang, Jian and Cheng, Ming-Ming},
  booktitle={Proceedings of the IEEE/CVF International Conference on Computer Vision},
  pages={18166--18176},
  year={2025}
}

@inproceedings{pan2025ice,
  title={Ice-bench: A unified and comprehensive benchmark for image creating and editing},
  author={Pan, Yulin and He, Xiangteng and Mao, Chaojie and Han, Zhen and Jiang, Zeyinzi and Zhang, Jingfeng and Liu, Yu},
  booktitle={Proceedings of the IEEE/CVF International Conference on Computer Vision},
  pages={16586--16596},
  year={2025}
}

@inproceedings{pathiraja2025refedit,
  title={RefEdit: A Benchmark and Method for Improving Instruction-based Image Editing Model on Referring Expressions},
  author={Pathiraja, Bimsara and Patel, Maitreya and Singh, Shivam and Yang, Yezhou and Baral, Chitta},
  booktitle={Proceedings of the IEEE/CVF International Conference on Computer Vision},
  pages={15646--15656},
  year={2025}
}

@inproceedings{gumulti,
  title={Multi-Reward as Condition for Instruction-based Image Editing},
  author={Gu, Xin and Li, Ming and Zhang, Libo and Chen, Fan and Wen, Longyin and Luo, Tiejian and Zhu, Sijie},
  booktitle={The Thirteenth International Conference on Learning Representations},
  year={2025}
}

@article{ghosh2023geneval,
  title={Geneval: An object-focused framework for evaluating text-to-image alignment},
  author={Ghosh, Dhruba and Hajishirzi, Hannaneh and Schmidt, Ludwig},
  journal={Advances in Neural Information Processing Systems},
  volume={36},
  pages={52132--52152},
  year={2023}
}

@article{wang2025gpt,
  title={Gpt-image-edit-1.5 m: A million-scale, gpt-generated image dataset},
  author={Wang, Yuhan and Yang, Siwei and Zhao, Bingchen and Zhang, Letian and Liu, Qing and Zhou, Yuyin and Xie, Cihang},
  journal={arXiv preprint arXiv:2507.21033},
  year={2025}
}

@article{tian2025mmada,
  title={MMaDA-Parallel: Multimodal Large Diffusion Language Models for Thinking-Aware Editing and Generation},
  author={Tian, Ye and Yang, Ling and Yang, Jiongfan and Wang, Anran and Tian, Yu and Zheng, Jiani and Wang, Haochen and Teng, Zhiyang and Wang, Zhuochen and Wang, Yinjie and others},
  journal={arXiv preprint arXiv:2511.09611},
  year={2025}
}

@inproceedings{nielarge,
  title={Large Language Diffusion Models},
  author={Nie, Shen and Zhu, Fengqi and You, Zebin and Zhang, Xiaolu and Ou, Jingyang and Hu, Jun and ZHOU, JUN and Lin, Yankai and Wen, Ji-Rong and Li, Chongxuan},
  booktitle={The Thirty-ninth Annual Conference on Neural Information Processing Systems},
  year={2025}
}

@inproceedings{pengdreambench++,
  title={DreamBench++: A Human-Aligned Benchmark for Personalized Image Generation},
  author={Peng, Yuang and Cui, Yuxin and Tang, Haomiao and Qi, Zekun and Dong, Runpei and Bai, Jing and Han, Chunrui and Ge, Zheng and Zhang, Xiangyu and Xia, Shu-Tao},
  booktitle={The Thirteenth International Conference on Learning Representations},
  year={2025}
}

@inproceedings{ruiz2023dreambooth,
  title={Dreambooth: Fine tuning text-to-image diffusion models for subject-driven generation},
  author={Ruiz, Nataniel and Li, Yuanzhen and Jampani, Varun and Pritch, Yael and Rubinstein, Michael and Aberman, Kfir},
  booktitle={Proceedings of the IEEE/CVF conference on computer vision and pattern recognition},
  pages={22500--22510},
  year={2023}
}

@article{li2023dreamedit,
  title={Dreamedit: Subject-driven image editing},
  author={Li, Tianle and Ku, Max and Wei, Cong and Chen, Wenhu},
  journal={arXiv preprint arXiv:2306.12624},
  year={2023}
}

@InProceedings{chang2022maskgit,
  title = {MaskGIT: Masked Generative Image Transformer},
  author={Huiwen Chang and Han Zhang and Lu Jiang and Ce Liu and William T. Freeman},
  booktitle = {The IEEE Conference on Computer Vision and Pattern Recognition (CVPR)},
  month = {June},
  year = {2022}
}

@inproceedings{tanghart,
  title={HART: Efficient Visual Generation with Hybrid Autoregressive Transformer},
  author={Tang, Haotian and Wu, Yecheng and Yang, Shang and Xie, Enze and Chen, Junsong and Chen, Junyu and Zhang, Zhuoyang and Cai, Han and Lu, Yao and Han, Song},
  booktitle={The Thirteenth International Conference on Learning Representations},
  year={2025}
}

@inproceedings{chang2023muse,
  title={Muse: Text-To-Image Generation via Masked Generative Transformers},
  author={Chang, Huiwen and Zhang, Han and Barber, Jarred and Maschinot, Aaron and Lezama, Jose and Jiang, Lu and Yang, Ming-Hsuan and Murphy, Kevin Patrick and Freeman, William T and Rubinstein, Michael and others},
  booktitle={International Conference on Machine Learning},
  pages={4055--4075},
  year={2023},
  organization={PMLR}
}

@article{team2024chameleon,
  title={Chameleon: Mixed-modal early-fusion foundation models},
  author={Team, Chameleon},
  journal={arXiv preprint arXiv:2405.09818},
  year={2024}
}

@article{comanici2025gemini,
  title={Gemini 2.5: Pushing the frontier with advanced reasoning, multimodality, long context, and next generation agentic capabilities},
  author={Comanici, Gheorghe and Bieber, Eric and Schaekermann, Mike and Pasupat, Ice and Sachdeva, Noveen and Dhillon, Inderjit and Blistein, Marcel and Ram, Ori and Zhang, Dan and Rosen, Evan and others},
  journal={arXiv preprint arXiv:2507.06261},
  year={2025}
}

@article{liang2025discrete,
  title={Discrete diffusion vla: Bringing discrete diffusion to action decoding in vision-language-action policies},
  author={Liang, Zhixuan and Li, Yizhuo and Yang, Tianshuo and Wu, Chengyue and Mao, Sitong and Nian, Tian and Pei, Liuao and Zhou, Shunbo and Yang, Xiaokang and Pang, Jiangmiao and others},
  journal={arXiv preprint arXiv:2508.20072},
  year={2025}
}

@article{liu2025longllada,
  title={Longllada: Unlocking long context capabilities in diffusion llms},
  author={Liu, Xiaoran and Song, Yuerong and Liu, Zhigeng and Huang, Zengfeng and Guo, Qipeng and He, Ziwei and Qiu, Xipeng},
  journal={arXiv preprint arXiv:2506.14429},
  year={2025}
}

@article{shi2025muddit,
  title={Muddit: Liberating generation beyond text-to-image with a unified discrete diffusion model},
  author={Shi, Qingyu and Bai, Jinbin and Zhao, Zhuoran and Chai, Wenhao and Yu, Kaidong and Wu, Jianzong and Song, Shuangyong and Tong, Yunhai and Li, Xiangtai and Li, Xuelong and others},
  journal={arXiv preprint arXiv:2505.23606},
  year={2025}
}

@inproceedings{zhang2025enabling,
  title={Enabling instructional image editing with in-context generation in large scale diffusion transformer},
  author={Zhang, Zechuan and Xie, Ji and Lu, Yu and Yang, Zongxin and Yang, Yi},
  booktitle={The Thirty-ninth Annual Conference on Neural Information Processing Systems},
  year={2025}
}

@article{lowe2004distinctive,
  title={Distinctive image features from scale-invariant keypoints},
  author={Lowe, David G},
  journal={International journal of computer vision},
  volume={60},
  number={2},
  pages={91--110},
  year={2004},
  publisher={Springer}
}

@inproceedings{muja2009fast,
  title={Fast approximate nearest neighbors with automatic algorithm configuration},
  author={Muja, Marius and Lowe, David G},
  booktitle={International conference on computer vision theory and applications},
  volume={1},
  pages={331--340},
  year={2009},
  organization={Scitepress}
}

\appendix
\newpage
\section{Theoretical Foundations of Grounding-Aware Discrete Inversion}

In this section, we provide a rigorous theoretical formulation of our Grounding-Aware Discrete Inversion framework. We cast the generation and inversion of Diffusion Large Language Models (DLLMs) within a formal probabilistic perspective, formulating the forward and reverse transitions as sequential stochastic sampling over discrete state spaces, driven by masked objective optimization and logit-level residual rectification.

\subsection{Preliminary: Probabilistic Formulation of Diffusion LLMs}

We first formalize the domain over which the diffusion process operates. Let a sequence of length $N$ be defined over a finite vocabulary of size $d$. 

\begin{definition}[Discrete State Space]
The discrete state space is defined as $\mathcal{D} = [d]^N$. The state variable at continuous time $t \in [0, 1]$ is denoted as $\mathbf{X}_t \in \mathcal{D}$, governed by a probability mass function $p_t(\mathbf{x})$.
\end{definition}

The training of DLLMs can be formulated as learning the transition dynamics within $\mathcal{D}$ via a masked token prediction objective. Let $\mathbf{X}_0 \sim q(\mathbf{x}_0)$ denote the empirical data distribution, and $\mathbf{X}_t$ be the corrupted state yielded by the forward process at a uniformly sampled time step $t \sim \mathcal{U}[0, 1]$.

\begin{definition}[Unified Diffusion Objective]
The parameterized model $p_\theta(\cdot | \mathbf{X}_t)$ is optimized by minimizing the expected negative log-likelihood over the masked tokens:
\begin{equation}
    \mathcal{L}_{\mathrm{unify}}(\theta) = -\mathbb{E}_{t, \mathbf{X}_0, \mathbf{X}_t} \left[ \frac{1}{t} \sum_{i=1}^N \mathbb{I}\left[\mathbf{X}_t^{(i)} = \mathrm{[MASK]}\right] \log p_\theta(\mathbf{X}_0^{(i)} | \mathbf{X}_t) \right],
\end{equation}
where $\mathbb{I}[\cdot]$ is the indicator function strictly evaluating on the corrupted token subset.
\end{definition}

Unlike continuous diffusion paradigms, generative processes in DLLMs require sampling from discrete categorical distributions over the vocabulary space during inference. Let $\Phi_\theta(\mathbf{X}_t, t) \in \mathbb{R}^d$ denote the unnormalized log-probabilities (logits) predicted by the network at step $t$.

To perform unbiased sampling from the categorical distribution parameterized by $\Phi_\theta$, we employ a temperature-scaled Gumbel-Max trick. We first introduce a standard uniform random variable $\mathbf{U} \sim \mathcal{U}(0, \mathbf{I})$ possessing the identical dimensional topology as the logit space.

\begin{proposition}[Temperature-Scaled Gumbel-Max Sampling]
By applying the probability integral transform, the independent and identically distributed Gumbel noise $\mathbf{G} \in \mathbb{R}^d$ is rigorously constructed from the uniform prior:
\begin{equation}
    \mathbf{G} = -\log(-\log(\mathbf{U} + \epsilon) + \epsilon), \quad \mathbf{U} \sim \mathcal{U}(0, \mathbf{I}),
\end{equation}
where $\epsilon \to 0^+$ serves as an infinitesimally small constant to ensure numerical stability against domain singularities. 

Given a temperature scaling factor $\tau > 0$, the discrete token transition for step $t-1$ is formally derived as a stochastic optimization problem over the vocabulary simplex:
\begin{equation}
    \mathbf{X}_{t-1} = \mathop{\arg\max}_{j \in \{1, \dots, d\}} \left( \frac{\Phi_\theta(\mathbf{X}_t, t)^{(j)}}{\tau} + \mathbf{G}^{(j)} \right).
\end{equation}
\end{proposition}

This formulation provides a strict bounding for the generation entropy. Specifically, as $\tau \to 0^+$, the stochastic noise $\mathbf{G}$ is completely suppressed, causing the transition to degenerate into a deterministic greedy search (i.e., absolute $\arg\max$). Conversely, $\tau > 0$ systematically controls the stochastic relaxation, preventing deterministic mode collapse and ensuring output diversity during the discrete generation processes.

\subsection{Theoretical Framework of Grounding-Aware Discrete Inversion}

Due to the non-deterministic nature of the discrete reverse process, exact inversion requires formulating structure-preserving priors constrained by a localized mapping. We define $\mathbf{M} \in \{0, 1\}^N$ as the binary spatial grounding mask.

To systematically reverse the generative process, the inversion trajectory must progressively corrupt the complete image by masking out the most predictable tokens. Let $\mathbf{P}_t \in (0, 1]^N$ be the predicted probability vector for the currently unmasked tokens at step $t$.

\begin{definition}[Stochastic Grounding-Aware Inversion Masking]
To prevent deterministic trajectory collapse during the backward masking process, we introduce a stochastic relaxation to the token confidence evaluation. The stochastic confidence field $\tilde{\mathbf{S}}_t \in \mathbb{R}^N$ is formulated by injecting temperature-scaled Gumbel noise into the log-probability space:
\begin{equation}
    \tilde{\mathbf{S}}_t = \log \mathbf{P}_t + \tau_{\mathrm{mask}} \cdot \mathbf{G}_{\mathrm{mask}}, \quad \mathbf{G}_{\mathrm{mask}} \sim \mathrm{Gumbel}(0, \mathbf{I}),
\end{equation}
where $\tau_{\mathrm{mask}} > 0$ dictates the magnitude of stochasticity, and probabilities are lower-bounded to avoid numerical singularities.

The masking capacity $N_t$, representing the exact quantity of tokens to be masked at time $t$, follows a non-linear sinusoidal schedule bounded by the spatial grounding prior $\mathbf{M}$:
\begin{equation}
    N_t = \left\lfloor \|\mathbf{M}\|_1 \cdot \sin\left(\frac{\pi t}{2T}\right) \right\rfloor.
\end{equation}

In the discrete inversion paradigm, the objective is to incrementally destruct the image information by aggressively masking the most structurally redundant tokens. To strictly localize this degradation, we derive a dynamic stochastic threshold $\eta_t$ using the $N_t$-th order statistic (specifically, the $N_t$-th largest element) of the stochastic confidence field within the grounded region:
\begin{equation}
    \eta_t = \mathrm{kth\_largest}\left( \left\{ \tilde{\mathbf{S}}_t^{(i)} \mid \mathbf{M}^{(i)} = 1 \right\}, N_t \right).
\end{equation}

Finally, the discrete binary masking indicator $\mathbf{m}_t \in \{0, 1\}^N$ for the current inversion step is formalized via the Heaviside step function $\Theta(\cdot)$. This operation deterministically masks the top-$N_t$ tokens possessing the highest stochastic confidence:
\begin{equation}
    \mathbf{m}_t^{(i)} = \mathbf{M}^{(i)} \cdot \Theta\left( \tilde{\mathbf{S}}_t^{(i)} - \eta_t \right).
\end{equation}
\end{definition}

To map the source structural priors into the discrete latent space without inducing rigid deterministic artifacts, we formulate the location-aware inversion residual $\mathbf{Z}_t$. Let $\mathbf{c}_{\mathrm{src}}$ and $\mathbf{c}_{\mathrm{tgt}}$ be the source and target textual conditionings, respectively. During the inversion phase, the denoiser predicts the probability landscape in the logit space conditioned on the source text:
\begin{equation}
    \hat{\mathbf{Y}}_t = \mathcal{D}_\theta(\mathbf{X}_t, \mathbf{c}_{\mathrm{src}}, t),
\end{equation}
where $\hat{\mathbf{Y}}_t \in \mathbb{R}^{N \times |\mathcal{V}|}$ represents the unnormalized pre-softmax predictions over the discrete vocabulary $\mathcal{V}$.

To guarantee that the exact reconstruction of the original image $\mathbf{X}_0$ is encoded within the discrete stochastic trajectory, we adapt the Location-Aware Argmax Inversion (LAI) \cite{dao2025discrete} to construct an oracle tensor $\mathbf{Y}_t$.

\begin{definition}[Location-Aware Argmax Inversion and Residual Extraction]
The LAI function \cite{dao2025discrete} explicitly rectifies the predicted logits using Gumbel truncation sampling to enforce precise token reconstruction, while preserving the original distributional shape for non-target labels. For the $i$-th token, let the predicted logit for the ground-truth token $\mathbf{X}_0^{(i)}$ be $l_{\max} = [\hat{\mathbf{Y}}_t]^{(i)}_{\mathbf{X}_0^{(i)}}$. We first sample a base value for the target token from the standard Gumbel distribution:
\begin{equation}
    q_{\max}^{(i)} \sim \mathrm{Gumbel}(\mu = l_{\max}, \beta = 1)
\end{equation}
For all other vocabulary indices $v \in \mathcal{V} \setminus \{\mathbf{X}_0^{(i)}\}$, let the predicted logit be $p = [\hat{\mathbf{Y}}_t]^{(i)}_v$. To ensure the probability of $\mathbf{X}_0^{(i)}$ remains strictly the largest, we sample from a truncated Gumbel distribution:
\begin{equation}
    q^{(i)}_v \sim \mathrm{GumbelTrunc}(\mu = p, \beta = 1, \mathrm{trunc} = q_{\max}^{(i)} - \tau)
\end{equation}
where $\tau > 0$ is a predefined margin. The $\mathrm{GumbelTrunc}$ sampling algorithm, given location $\phi$ and threshold $T$, is formulated as $\phi - \log(\exp(\phi - T) - \log u)$ with $u \sim \mathrm{Uniform}(0, 1)$, effectively subtracting a dynamic penalty to yield a strictly bounded smaller value.

The oracle tensor $\mathbf{Y}_t = \mathrm{LAI}(\mathbf{X}_t, \hat{\mathbf{Y}}_t, \mathbf{X}_0)$ is thus constructed as:
\begin{equation}
    [\mathbf{Y}_t]^{(i)}_{v} = 
    \begin{cases} 
        q_{\max}^{(i)}, & \text{if } v = \mathbf{X}_0^{(i)} \\ 
        q^{(i)}_v, & \text{otherwise} 
    \end{cases}
\end{equation}

The structural inversion residual $\mathbf{Z}_t$, which mathematically encodes the exact momentum discrepancy required to anchor the discrete trajectory to $\mathbf{X}_0$, is extracted as the direct differential:
\begin{equation}
    \mathbf{Z}_t = \mathbf{Y}_t - \hat{\mathbf{Y}}_t.
\end{equation}
\end{definition}

During the editing (reverse generation) stage, the target prompt $\mathbf{c}_{\mathrm{tgt}}$ induces a semantically shifted logit distribution $\hat{\mathbf{Y}}'_t = \mathcal{D}_\theta(\mathbf{X}_t, \mathbf{c}_{\mathrm{tgt}}, t)$.

\begin{figure}[tb]
  \centering
  \includegraphics[width=\linewidth]{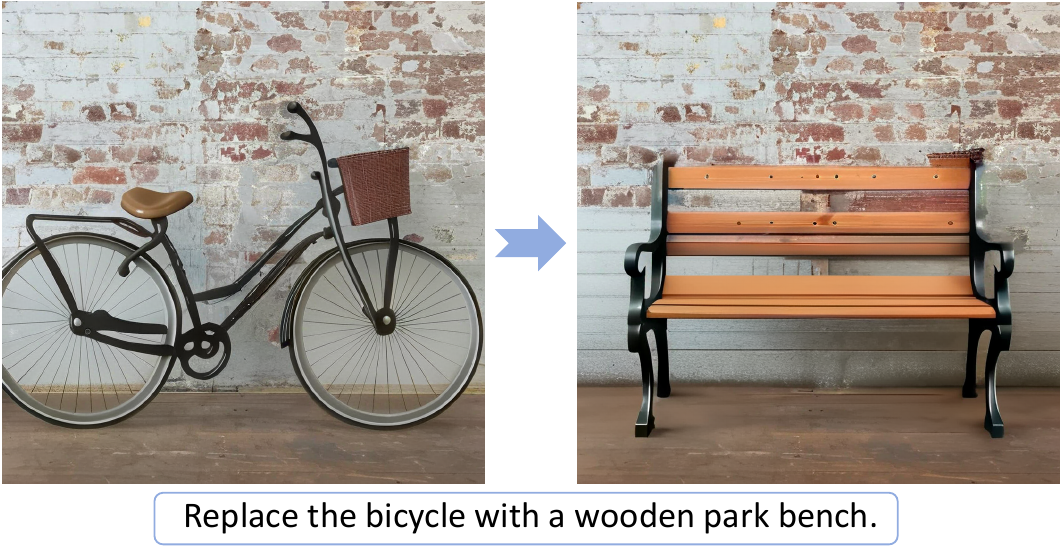}
    \caption{An example of Mask Relaxation. Expanding the tight source mask $\mathbf{M}_{\text{src}}$ to its bounding box accommodates noticeable shape variations (e.g., bicycle to bench), ensuring complete object generation while preserving the unedited background.}
  \label{fig:appendix_mask}
\end{figure}

\begin{proposition}[Stochastic Logit Rectification and Residual Injection]
Given a diagonal mixing matrix $\mathbf{\Lambda} = \mathrm{diag}(\lambda_1, \dots, \lambda_N)$ with $\lambda_i \in [0, 1]$, the rectified logit $\tilde{\mathbf{Y}}_t$ and the final edited discrete transition are formulated as:
\begin{equation}
    \tilde{\mathbf{Y}}_t = \hat{\mathbf{Y}}'_t + \mathbf{\Lambda} \mathbf{Z}_t,
\end{equation}
\begin{equation}
    \mathbf{X}_{t-1}^{\mathrm{edit}} = \mathop{\arg\max} \left( \tilde{\mathbf{Y}}_t + \gamma \mathbf{G} \right), \quad \mathbf{G} \sim \mathrm{Gumbel}(0, \mathbf{I}),
\end{equation}
where $\gamma$ controls the stochastic relaxation.
\end{proposition}

This formulation dynamically interpolates between target semantic editability and source structural fidelity, preserving the requisite entropy for high-frequency detail generation.

\section{Mask Relaxation for Shape Variations}
\label{sec:appendix_mask}

As detailed in \cref{sec3.3:refinement_module}, the Residual Recovery mechanism in \ourmethod{} is highly robust for general editing tasks. By operating on the precise residual region $\mathbf{M}_{\text{res}} = \mathbf{M}_{\text{src}} \setminus \mathbf{M}_{\text{tgt}}$, it efficiently isolates and restores the exposed background with exceptional efficacy. While this tight, pixel-level mask ($\mathbf{M}_{\text{src}}$) ensures seamless blending, it can occasionally constrain the generative process when the target entity exhibits a different geometric structure or necessitates a larger spatial footprint, potentially leading to minor vestigial artifacts.

To gracefully resolve these spatial differences while strictly preserving the unedited regions, we introduce a flexible mask relaxation strategy. Specifically, we redefine $\mathbf{M}_{\text{src}}$ by computing a bounding box that fully encompasses the original object. This spatial relaxation gives \ourmethod{} ample flexibility to handle structural differences and synthesize larger objects, while keeping the background outside the bounding box entirely unchanged.

As visually demonstrated in \cref{fig:appendix_mask}, replacing a bicycle with a wooden park bench presents a noticeable structural difference. By expanding the $\mathbf{M}_{\text{src}}$ of the bicycle to its encompassing bounding box, our method effectively mitigates spatial conflicts. This expanded context grants the model sufficient spatial freedom to fully render the bench's broader structure, preventing structural blending artifacts and culminating in a highly realistic and visually coherent editing result.

\begin{figure}[tb]
  \centering
  \includegraphics[width=\linewidth]{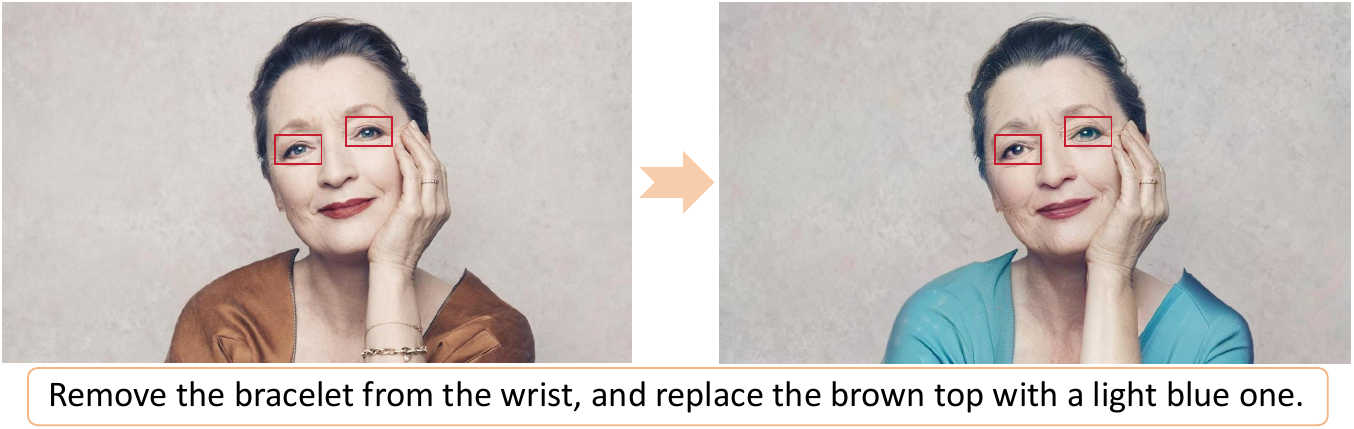}
    \caption{An example of VQModel reconstruction characteristics. \ourmethod{} executes semantic edits accurately, but the VQModel's decompression introduces minor color variations in high-frequency details like the eyes. Zoom in for better view.}
  \label{fig:appendix_vqloss}
\end{figure}

\section{Analysis of Reconstruction Artifacts}

As discussed in \cref{5.2.2:Preservation of Non-edited Regions.}, \ourmethod{} excels at executing complex semantic instructions while preserving overall visual content. However, we observe minor variations in low-level precision compared to specialized pixel-level methods. This behavior is primarily associated with the reconstruction phase of the VQModel within the Lumina-DiMOO backbone during image decompression. 

\cref{fig:appendix_vqloss} provides a visual example of this phenomenon. Given the instruction to remove the bracelet and change the top color to light blue, \ourmethod{} successfully and cleanly fulfills the primary semantic objectives. Upon closer inspection of high-frequency regions, such as the subject's eyes, subtle texture and color deviations become visible. Specifically, the eye on the left transitions from its original blue color to a pattern featuring a blue center surrounded by brown. Meanwhile, the eye on the right develops a distinct blue tint in the upper left area of the sclera and along the eyelashes. 

These gentle variations illustrate how the current VQModel handles fine-grained details during the image compression and decompression cycle. Importantly, the semantic control achieved by \ourmethod{} remains highly precise. We view these minor artifacts as a temporary characteristic of the underlying architecture, which will improve with the evolution of more robust autoencoder models.

\section{Implementation Details}

In this section, we provide supplementary details to smoothly facilitate the reproduction of our experiments. Regarding the grounding module introduced in \cref{sec3.1:grounding_module}, we flexibly adapt the segmentation foundation model according to the input format. Specifically, we utilize SAM 3~\cite{carion2025sam} to elegantly process textual and box inputs, while SAM 2~\cite{ravisam} is carefully adopted to handle point inputs.

Furthermore, \cref{tab:appendix_generating_para} outlines the chosen generation setups for the training-free and training-based methods evaluated in our study. We believe these optimal configurations allow \ourmethod{} to properly demonstrate its robust capabilities and consistent advantages. For any remaining baseline methods or hyperparameters not explicitly listed, we naturally adopt their standard default values.

\begin{table}[th!]
\centering
\begin{tabular}{p{0.3\linewidth} | p{0.6\linewidth}}
\toprule
\textbf{Method} & \textbf{Generation Setup} \\
\midrule

\ourmethod{}+Lumina-DiMOO & cfg\_scale = 4.0, timesteps = 64, temperature = 1.0\\
\midrule
\ourmethod{}+MMaDA & cfg\_scale = 4.0, timesteps = 64, temperature = 1.0\\
\midrule
DICE+Lumina-DiMOO & cfg\_scale = 4.0, timesteps = 64, temperature = 1.0\\
\midrule
DirectInversion+PnP & guidance\_scale = 7.5, num\_steps = 50 \\
\midrule
DirectInversion+P2P & guidance\_scale = 7.5, cross\_replace\_steps = 0.4, self\_replace\_steps = 0.6 \\
\midrule
Edit-R1 & true\_cfg\_scale = 4.0, num\_inference\_steps = 40\\
\midrule 
LongCat & guidance\_scale = 4.5, num\_inference\_steps = 50\\
\midrule
Qwen-Image & true\_cfg\_scale = 4.0, num\_inference\_steps = 50\\
\midrule
FLUX.1-Kontext & guidance\_scale = 2.5\\
\midrule
OmniGen2 & text\_guidance\_scale = 5.0, image\_guidance\_scale = 2.0, num\_inference\_step = 50\\
\midrule
Lumina-DiMOO & cfg\_scale = 2.5, cfg\_img = 4.0, timesteps = 64, temperature = 1.0 \\
\midrule
OneDiffusion &  guidance\_scale = 4, num\_inference\_steps = 50 \\

\bottomrule
\end{tabular}
\vspace{1ex}
\caption{Generating parameters for training-free and training-based methods.}
\label{tab:appendix_generating_para}
\end{table}

\end{document}